\documentclass{article} 
\usepackage{iclr2027_conference,times}

\usepackage{amsmath,amsfonts,bm}

\def\eqref#1{equation~\ref{#1}}

\def\1{\bm{1}}

\DeclareMathAlphabet{\mathsfit}{\encodingdefault}{\sfdefault}{m}{sl}
\SetMathAlphabet{\mathsfit}{bold}{\encodingdefault}{\sfdefault}{bx}{n}

\usepackage[breaklinks=true]{hyperref}
\usepackage{url}
\usepackage{amsmath,amssymb,amsfonts,amsthm}
\usepackage{bm}
\usepackage{nicefrac}
\usepackage{booktabs}
\usepackage{multirow}
\usepackage{xcolor}
\usepackage{colortbl}
\usepackage{enumitem}
\usepackage{pifont}
\usepackage{graphicx}
\usepackage[font=small]{caption}
\usepackage{subcaption}
\hypersetup{colorlinks=true,
            linkcolor=blue!55!black,
            citecolor=green!40!black,
            urlcolor=blue!55!black}
\newtheorem{proposition}{Proposition}

\newcommand{\x}{\bm{x}}                       
\newcommand{\Pset}{\mathcal{P}}               
\newcommand{\np}{M}                           
\newcommand{\coal}{S}                         
\newcommand{\fillx}[1]{\tilde{\x}_{#1}}       
\newcommand{\clf}{f}                          
\newcommand{\val}{v}                          
\newcommand{\shap}{\phi}                      
\newcommand{\shaphat}{\hat{\phi}}             
\newcommand{\gtstd}{\shap^{\star\mathrm{-mar}}}   
\newcommand{\gtman}{\shap^{\star\mathrm{-cond}}}  
\newcommand{\ECone}{\mathrm{EC1}}
\newcommand{\ECthree}{\mathrm{EC3}}
\newcommand{\msd}[1]{{\scriptsize$\pm$#1}}    

\title{\flushleft MotionBench-XAI: Benchmarking \\ What Shapes Shapley Explanations in Spatiotemporal Machine Learning}

\author{Mathilde Papillon \\
Department of Physics\\
University of California, Santa Barbara\\
\texttt{papillon@ucsb.edu}
\And
Louisa Cornelis \\
Department of Physics\\
University of California, Santa Barbara\\
\And
Nina Miolane \\
Department of Electrical and Computer Engineering\\
University of California, Santa Barbara\\
}

\iclrfinalcopy
\begin{document}

\maketitle
\lhead{ } 

\begin{abstract}
Predicting outcomes from spatiotemporal data, such as assessing Parkinson's severity from human motion, often requires explainability in addition to accuracy. Practitioners need to know what drove the prediction, whether it be spatial (a body joint), temporal (a specific window), or both. \textit{Shapley values} are attributions computed by treating input features as players in a cooperative game and measuring how the model output changes as coalitions of players are removed. Computing them involves choosing what counts as players, how absent players are imputed, and how the game is evaluated. How these choices affect explanation quality is poorly understood for spatiotemporal data, and real data carries no ground-truth attributions to check against. 
We present \textbf{MotionBench-XAI}, a benchmark built on i) five synthetic spatiotemporal datasets with exact ground-truth Shapley values, ii) three real datasets across clinical gait, electrocardiogram, and audio, iii) ten attribution methods across three player sets, and iv) three trained architectures.
We show that grouping features into players, across time, space, or both, matters more than the choice of imputer. Generative imputers improve explanations only when their conditional models are accurate. On real data, the standard proxy metrics select contradictory best methods. We distill these findings into practical insights and provide an open-source pipeline.
\end{abstract}

\section{Introduction}
\label{sec:intro}

Spatiotemporal data, such as 3D motion skeletons, multi-lead electrocardiograms, and audio signals, are used in machine learning for safety-critical domains including clinical diagnostics and monitoring. In these settings, practitioners often demand explainability alongside predictive accuracy. Beyond knowing that a prediction is accurate, they wish to understand which spatial components of a sequence, and at which times, drove that prediction.
Shapley-value attribution \citep{shapley1953value,lundberg2017unified} is
one of the most widely deployed tools to quantify the importance of input features in a model's final prediction
\citep{arrieta2020explainable,slack2020fooling}. For example, it quantifies which joints in a patient's skeleton drive a Parkinson's severity prediction. The method treats model inputs as \textit{players} in a cooperative \textit{game} and computes the contribution, called Shapley value, of each player to the prediction.

Computing Shapley values requires computing the model's updated prediction when it is only shown a subset of players, for many different subsets. Since standard models cannot accept missing inputs, absent features are imputed using a variety of methods. Two active lines of work adapt the classical Shapley method to structured temporal data. \emph{Manifold-aware} variants change the imputer, restricting imputed players to the data manifold
\citep{frye2021shapley,aas2021explaining,olsen2022using,taufiq2023manifold, zhang2026axiomatic}.  \emph{Temporal} variants organize the players in terms of time steps, and thus report attributions on a temporal scale~\citep{bento2021timeshap,NAYEBI2023104438,villani2022feature,
de2025shats}.
Each of those Shapley variants makes a series of design choices in the process of producing an explanation. The first choice defines what counts as players (whole time windows, whole joints, or fine joint$\times$window cells). The second choice defines how masked players are replaced (zeros, averages, copies of other sequences, or samples from a generative model of the data manifold). The third choice decides how the cooperative game is played, and what the ground truth, if it existed, would even be. The final choice is the \emph{metric} used to grade the quality of the explanation, with or without access to that ground truth.
%
%
These design choices, while critical to the outcome, are, to the best of our knowledge, never benchmarked jointly. This is especially critical for spatiotemporal data, which exhibits both feature dependence across channels and temporal coherence.

\begin{figure}[t]
  \centering
  \includegraphics[width=1.0\linewidth]{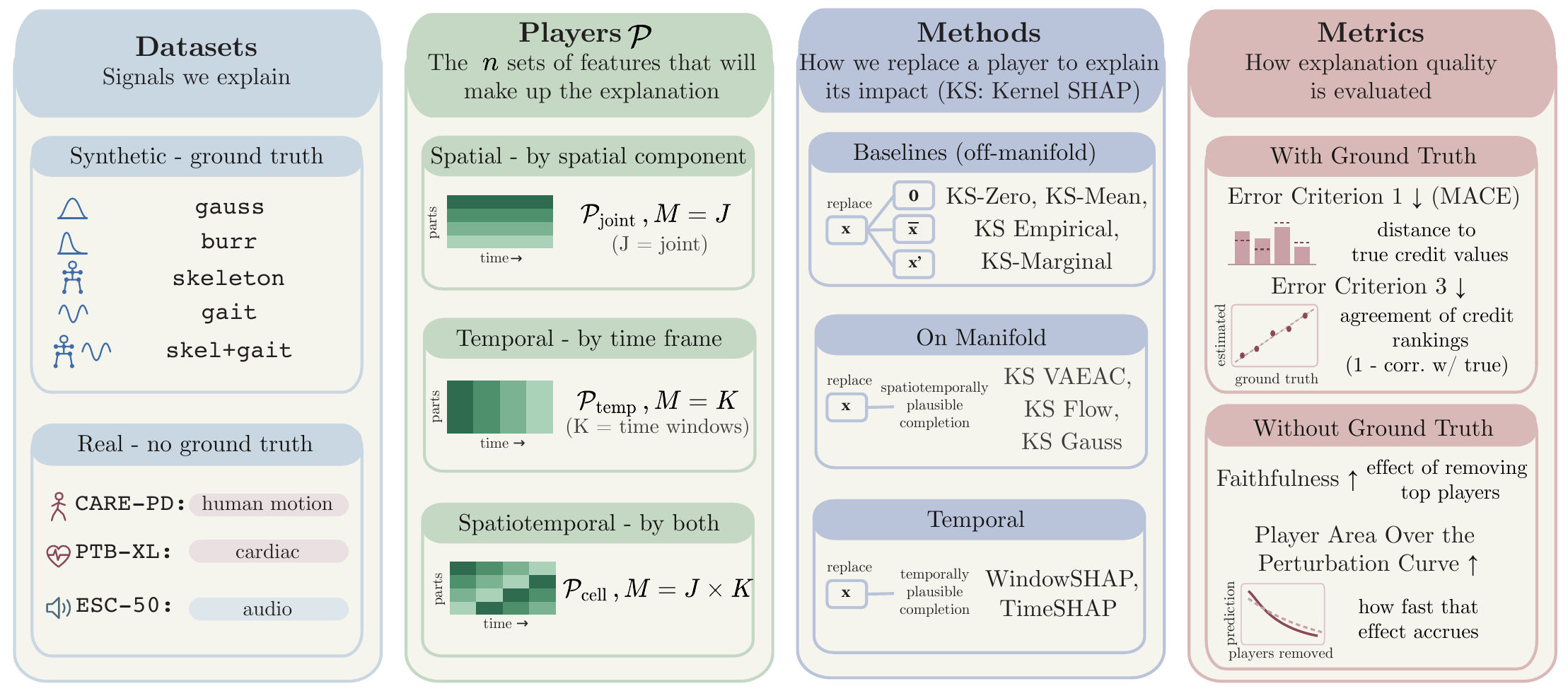}
  \caption{\textbf{MotionBench-XAI} evaluates spatiotemporal Shapley value attributions as a feature-based explainability method. 
\textit{Datasets (first column):} Five synthetic spatiotemporal datasets paired with ground-truth Shapley values, and three real-world datasets. 
\textit{Players $\mathcal{P}$ (second column):} Input features partitioned spatially ($P_{\text{joint}}$), temporally ($P_{\text{temp}}$), or across a spatiotemporal grid ($P_{\text{cell}}$). 
\textit{Imputation methods (third column):} Strategies for replacing absent players via off-manifold baselines, on-manifold generative models, or temporal SHAP variants. 
\textit{Metrics (fourth column):} Evaluation via exact attribution error and linear agreement error for synthetic cases, or proxy metrics (faithfulness, PlayerAOPC) for real-world cases where ground truth is absent.}
  \label{fig:overview}
\end{figure}

\paragraph{Contributions.}
To remedy this, we propose \textbf{MotionBench-XAI}, a benchmark to compare on and off-manifold Shapley methods on data with both feature dependence and temporal coherence (Figure~\ref{fig:overview}).  In both synthetic and real-world settings, MotionBench-XAI serves as an evaluation instrument to measure how hidden design choices dictate explanation quality for spatiotemporal data.
\begin{itemize}[leftmargin=*]
\item We construct five spatiotemporal dataset families from varied distributions with closed-form marginal and conditional expectations, allowing ground-truth Shapley values  per trained classifier (\S\ref{sec:datasets-synth}, \S\ref{sec:bg-gt}).
\item We perform a systematic sweep of every combination of player set (temporal, spatial, spatiotemporal), imputation family (standard, temporal, generative), ground truth, and metric, per dataset and per classifier, using one explanation pipeline (\S\ref{sec:results-synth}).
\item We run the same pipeline on three real datasets where
  ground truth is not available (\S\ref{sec:datasets-real}): clinical gait (CARE-PD;
  \citealp{adeli2025multi}), 12-lead ECG (PTB-XL;
  \citealp{wagner2020ptbxl}), and environmental audio (ESC-50;
  \citealp{piczak2015esc}) (\S\ref{sec:results-real}).
\item We summarize findings with direct practical consequence for practitioners using Shapley explanations in applied settings, distilled into a practical guide (Fig. \ref{fig:decision}).
\end{itemize}

\S\ref{sec:related} situates MotionBench-XAI within the literature detailing how it resolves current evaluation gaps. All code and  checkpoints are released at \href{https://github.com/mathildepapillon/motionbench-xai}{github.com/mathildepapillon/motionbench-xai} (Appendix~\ref{app:repro}).

\section{Background and Methods}
\label{sec:background}

We introduce Shapley explanation and the notation for the four choices underlying a given explanation (Figure~\ref{fig:overview}): the \emph{players} (\S\ref{sec:bg-players}), the \emph{imputation} (\S\ref{sec:bg-imputation}), the \emph{ground truth} (\S\ref{sec:bg-gt}), and the \emph{metric} (\S\ref{sec:bg-metrics}). MotionBench-XAI varies each choice independently. A glossary in Table~\ref{tab:notation} lists the full notation.

\subsection{Shapley values on spatiotemporal data}
\label{sec:bg-shapley}

A spatiotemporal instance is a tensor $\x \in \mathbb{R}^{J \times F \times T}$:
$J$ spatial units (skeleton joints, ECG leads, frequency bands), $F$ coordinates per unit (e.g.\ $F{=}3$ for $x,y,z$ positions), and $T$ time frames.  A trained classifier $\clf : \mathbb{R}^{J \times F \times T} \to \mathbb{R}$ maps $\x$ to a prediction, and the goal of an attribution is
to divide attribution for that prediction among parts of $\x$.

Shapley attribution \citep{lundberg2017unified} formalizes this as a cooperative game. A \emph{player set} $\Pset = \{1, \dots, \np\}$ partitions the entries of $\x$ into $\np$ disjoint groups (players).  For a coalition of players $\coal \subseteq \Pset$, the \emph{value function}, $\val$, evaluates the model on the instance with the players outside $\coal$ hidden:
\begin{equation}
  \val(\coal) \;=\; \clf\!\left(\fillx{\coal}\right),
  \label{eq:value}
\end{equation}
where $\fillx{\coal}$ denotes $\x$ with the entries of players in $\coal$ kept and all other entries replaced by an \emph{imputation} (defined in \S\ref{sec:bg-imputation}). We write $\bar{\coal} = \Pset \setminus \coal$ for the hidden players, $\x_{\coal}$ for the visible entries, and $\x_{\bar{\coal}}$ for the hidden block, so that $\fillx{\coal}$ is assembled from $\x_{\coal}$ and an imputation $\tilde{\x}_{\bar{\coal}}$ of the rest.
When the imputer is random, MotionBench-XAI draws $R$ imputations and averages them into one completion, $\tilde{\x}_{\bar{\coal}} = \frac{1}{R} \sum_{r=1}^{R} \tilde{\x}^{[r]}_{\bar{\coal}}$. Hence, we evaluate the classifier only once on the averaged fill $\tilde{\x}_{\bar{\coal}}$. Every stochastic imputer uses the same draw count, $R{=}5$ on the synthetic suite and $R{=}1$ on the real datasets (Appendix~\ref{app:coalition_protocol}). Every game value is therefore reproducible. The ground truths of
\S\ref{sec:bg-gt} replace the averaged draws with the exact conditional or
marginal mean, the $R \to \infty$ limit of this average.


The Shapley value \citep{shapley1953value} of player $i$ is its average marginal contribution over all possible coalitions of players in $\Pset$,
\begin{equation}
  \shap_i \;=\; \sum_{\coal \subseteq \Pset \setminus \{i\}}
  \frac{|\coal|!\,(\np - |\coal| - 1)!}{\np!}
  \left[\val(\coal \cup \{i\}) - \val(\coal)\right].
  \label{eq:shapley}
\end{equation}

We write $\shap = (\shap_1, \dots, \shap_{\np})^{\!\top} \in \mathbb{R}^{\np}$ for the vector of Shapley values across all $\np$ players. Exact enumeration is exponential in $\np$. All methods in this paper estimate Eq.~\ref{eq:shapley} with KernelSHAP \citep{lundberg2017unified}, a weighted least-squares regression over sampled coalitions. We write $\shaphat \in \mathbb{R}^{\np}$ for the KernelSHAP estimate. We hold this estimator fixed across all KernelSHAP methods, evaluating the same number of coalitions throughout (Appendix~\ref{app:coalition_protocol}).
The temporal SHAP baselines run their reference implementations
(\S\ref{sec:bg-imputation}), so their rows also reflect their own estimators.

\subsection{What can a player be?}
\label{sec:bg-players}

\begin{figure}[t]
  \centering
  \includegraphics[width=\linewidth]{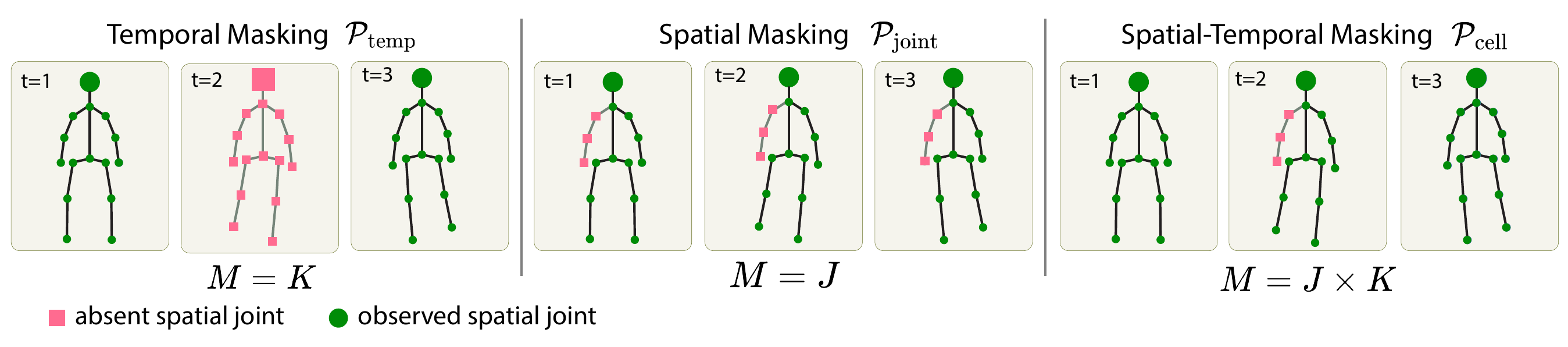}
  \caption{\textbf{What can a player be?}  The three player sets, shown as
  masking on a skeleton sequence: a time window ($\Pset_{\mathrm{temp}}$), a spatial joint ($\Pset_{\mathrm{joint}}$), or a joint$\times$window cell
  ($\Pset_{\mathrm{cell}}$).}
  \label{fig:players}
\end{figure}

The player set $\Pset$ decides how the input features will be preemptively divided before attributions are assigned (Figure~\ref{fig:overview}, second column). Figure~\ref{fig:players} shows the three families of players MotionBench-XAI evaluates on an example skeleton sequence. In every family, a player always includes all $F$ coordinates of the entries it covers. \textbf{Temporal
players} $\Pset_{\mathrm{temp}}$ cut the $T$ frames into $K$ uniform windows, and each window is one player ($\np = K$): this is the player set of temporal SHAP methods. \textbf{Spatial players} $\Pset_{\mathrm{joint}}$ make each spatial unit one player ($\np = J$). \textbf{Spatiotemporal players}
$\Pset_{\mathrm{cell}}$ combine the two, one player per joint $\times$ window cell ($\np = J \times K$). For example, on the CARE-PD gait data, the same
instance has $K=4$ windows, $J=17$ joints, which gives $M=68$ spatiotemporal players, also called cells.

\subsection{What can an imputation be?}
\label{sec:bg-imputation}

Once we have our players, Eq.~\ref{eq:value} requires masking some of these players turn by turn. To avoid having to modify the model being explained, the masked players need to be imputed by some other value. Writing $x_i$ for the entries of a single hidden player $i \in \bar{\coal}$, MotionBench-XAI considers three families of methods (Figure~\ref{fig:overview}, third column). 

\textbf{Standard (off manifold).}  The hidden player is replaced by a value that ignores the rest of the sequence: zeros ($x_i = 0$), the training mean
($x_i = \mathbb{E}[x_i]$), or a draw from the marginal ($x_i \sim P(x_i)$,
copied from another sequence in the evaluation pool). These define the
methods KS-Zero and KS-Mean \citep{lundberg2017unified}, KS-Marginal \citep{aas2021explaining}, and KS-Empirical \citep{aas2021explaining,redelmeier2020explaining}. KS-Empirical refines KS-Marginal by copying values from the sequences most similar to the explained one, weighted by a kernel. These imputations are cheap and model-agnostic but off manifold \citep{frye2021shapley}. On human skeleton data, for example, they can produce anatomically impossible motion.

\textbf{Temporal (on the temporal manifold).} WindowSHAP
\citep{NAYEBI2023104438} and TimeSHAP \citep{bento2021timeshap} keep the
standard zero fill. They distinguish themselves by cleverly picking temporal players to begin with: fixed windows (WindowSHAP-S), adaptively split windows (WindowSHAP-D), or per-event scoring with pruning (TimeSHAP). We evaluate them at temporal players only, aggregating per-timestep attributions into the $K$ uniform windows.

\textbf{Generative (on manifold).}  This family uses a conditional model to impute the hidden player from the visible ones, $x_i \sim P(x_i \mid \x_{\coal})$. KS-VAEAC (\citealp{olsen2022using,ivanov2018variational}) and KS-Flow use learned conditional models; their formulations are given in
Appendix~\ref{app:imputer_detail}.
We build KS-Flow for this benchmark so that the generative family contains two learned imputers of different strength. It mirrors KS-VAEAC except that it replaces the VAEAC generative model with a flow matching model \citep{lipman2023flow}. KS-Gauss (\citealp{aas2021explaining}) is the closed-form Gaussian conditional, with no learned components. These imputers stay near the data manifold but are learned or assumption-bound, hence imperfect. 

\subsection{What can a ground truth be?}
\label{sec:bg-gt}
In cooperative game theory, a game is defined as a rule that assigns a score to every coalition of players. In our setting, that rule is Eq.~\ref{eq:value}. The score of a coalition is the model's prediction on the input with the remaining players imputed, so the imputation is part of the rule. When we change how the hidden players are filled, the same
coalition receives a different score. We are then playing a different game, and Eq.~\ref{eq:shapley} assigns different attributions to the same input. In this work, two games stand out. If we fill the hidden players with their marginal expectation $\mathbb{E}[\x_{\bar{\coal}}]$, we play the \emph{marginal game}. If we instead fill them with their conditional expectation $\mathbb{E}[\x_{\bar{\coal}} \mid \x_{\coal}]$, we play the \emph{conditional game}~\citep{janzing2020feature,chen2020true}. The standard and temporal imputers of \S\ref{sec:bg-imputation} approximate the marginal game. The generative imputers approximate the conditional game. A \textit{ground truth Shapley value} solves a game exactly, using the expectation itself instead of an estimate of it. 


On synthetic data drawn from a well-chosen distribution, both expectations are available in closed form for every coalition. We can therefore compute both ground truths exactly (see App.~\ref{app:oracle_proofs}). For player sets with at most 12 players we enumerate every coalition, so the ground-truth Shapley values carry no error at all. For larger player sets we solve the same exact games on large sampled coalition designs and measure the remaining target error directly. Importantly, since the two exact imputations differ, the two ground truths assign different attributions. MotionBench-XAI grades each imputer method against the ground truth of its own game, which we call \emph{game-matched} grading. Where possible, we also report the standard convention of grading everything against $\gtman$. On real data, neither expectation is knowable. 


\subsection{What metric do we use to score the quality of an explanation?}
\label{sec:bg-metrics}

\paragraph{With ground truth (synthetic data).} Following
\citet{aas2021explaining,olsen2022using}, MotionBench-XAI reports two error criteria (EC, following \citet{olsen2022using}):
\begin{equation}
  \ECone = \tfrac{1}{\np}\,\big\lVert \shaphat - \shap^{\star} \big\rVert_1
  \qquad\text{and}\qquad
  \ECthree = 1 - r\!\left(\shaphat,\, \shap^{\star}\right),
\end{equation}
where $\shap^{\star}$ is the game-matched ground truth and $r$ is the Pearson
correlation per sequence. $\ECone$ is the mean absolute attribution error.
$\ECthree$ is its scale-invariant counterpart that measures how well the player attributions' order and weight match the ground truth. Lower is better for both.

\paragraph{Without ground truth (real data).} MotionBench-XAI uses the two standard deletion-based tests. \emph{Faithfulness correlation} \citep{bhatt2020evaluating} correlates the summed attributions of hidden players with the model's performance drop across coalitions. \emph{PlayerAOPC} \citep{samek2017evaluating} removes players in decreasing-attribution order and records the model's response. Higher is better for both. For both metrics, $\val(\coal)$ is computed with each method's own imputer, on the same coalitions. $\ECthree$ and faithfulness can be compared across player sets. However, $\ECone$ and PlayerAOPC scale with the number of players, so their values compare only within one player set.

\section{Related Work}
\label{sec:related}

\begin{table}[t]
\centering
\caption{\textbf{Coverage of MotionBench-XAI's evaluation dimensions.}
Synthetic ground truth provides known informative features or numerical
attribution targets. Conditional imputers fill missing inputs using the
observed inputs and are evaluated within explanation methods.
Multiple player sets means comparing different groupings of input features.
$\checkmark$ indicates a reported component and -- an unreported one.}
\label{tab:related_bench}
\setlength{\tabcolsep}{3pt}
\scriptsize
\begin{tabular*}{\linewidth}{@{\extracolsep{\fill}}lcccc@{}}
\toprule
Study
& \shortstack{Synthetic ground truth}
& \shortstack{Temporal data}
& \shortstack{Conditional imputers}
& \shortstack{Multiple player sets} \\
\midrule
BAM~\citep{Yang2019BenchmarkingAM}
 & $\checkmark$ & -- & -- & -- \\
CLEVR-XAI~\citep{arras2022clevr}
 & $\checkmark$ & -- & -- & -- \\
\citet{olsen2024comparative}
 & $\checkmark$ & -- & $\checkmark$ & -- \\
XTSC-Bench~\citep{hollig2023xtsc}
 & $\checkmark$ & $\checkmark$ & -- & -- \\
xaitimesynth~\citep{baer2026xaitimesynth}
 & $\checkmark$ & $\checkmark$ & -- & -- \\
SHAPformer~\citep{hertel2026shapformer}
 & $\checkmark$ & $\checkmark$ & -- & $\checkmark$ \\
TimeSHAP~\citep{bento2021timeshap}
 & -- & $\checkmark$ & -- & $\checkmark$ \\
ShaTS~\citep{de2025shats}
 & -- & $\checkmark$ & -- & $\checkmark$ \\
GroupSegment-SHAP~\citep{kim2026groupsegment}
 & $\checkmark$ & $\checkmark$ & -- & $\checkmark$ \\
\midrule
\textbf{MotionBench-XAI (ours)}
 & $\checkmark$ & $\checkmark$ & $\checkmark$ & $\checkmark$ \\
\bottomrule
\end{tabular*}
\end{table}

Prior work addresses dependence-aware Shapley attribution, temporal player
definitions, and explanation evaluation. MotionBench-XAI studies how these
choices interact (Table~\ref{tab:related_bench}).

\paragraph{Dependence-aware and temporal Shapley attribution.}
Marginal-background KernelSHAP~\citep{lundberg2017unified} fills missing
features with background values without conditioning on observed features.
With dependent features, this can produce off-manifold
inputs~\citep{frye2021shapley}. Conditional Shapley methods include
empirical and copula-based conditioning~\citep{aas2021explaining,aas2021vineexplaining}
and conditional inference trees~\citep{redelmeier2020explaining}.
Variational approaches~\citep{olsen2022using} use
VAEAC~\citep{ivanov2018variational} to sample missing features.
ManifoldShap restricts interventional evaluations to a specified domain
of validity~\citep{taufiq2023manifold}. Related work studies transport-based
paths for Aumann--Shapley attribution~\citep{zhang2026axiomatic}.

For sequences, TimeSHAP provides feature, timestep and cell
attributions~\citep{bento2021timeshap}, while WindowSHAP groups adjacent
time points within each variable~\citep{NAYEBI2023104438}.
\citet{villani2022feature} develop autoregressive surrogates and
time-consistent attributions. ShaTS studies temporal and feature
groupings~\citep{de2025shats}, using background substitution.
GroupSegment-SHAP combines dependence-based variable groups with temporal
segments~\citep{kim2026groupsegment} and mainly uses mean replacement.
Grouping alone does not provide conditional completion.

Marginal and conditional Shapley games answer different explanatory
questions~\citep{janzing2020feature,chen2020true,sundararajan2020many,kumar2020problems}.
Our synthetic targets use marginal or conditional mean fills
(\S\ref{sec:bg-gt}). Predicting on these fills generally differs from
averaging predictions over imputations.
Amortized estimators~\citep{jethani2022fastshap} and improved
regression-based estimation~\citep{covert2021improving} address
computational efficiency.

\paragraph{Explanation evaluation.}
Fixed-model deletion tests can confound feature importance with the
distribution shift caused by replacing inputs. ROAR addresses this by
retraining on modified data~\citep{hooker2019roar}.
\citet{bhatt2020evaluating} study faithfulness and sensitivity, while
Quantus provides evaluation metrics and illustrates how their settings
affect rankings~\citep{hedstrom2023quantus}.

\paragraph{Benchmarks and spatiotemporal applications.}
Synthetic time-series benchmarks identify informative
regions~\citep{ismail2020benchmarking,hollig2023xtsc,baer2026xaitimesynth}.
These masks differ from numerical Shapley values
(Table~\ref{tab:related_bench}).
\citet{olsen2022using,olsen2024comparative} evaluate conditional Shapley
estimates in tabular simulations with known distributions.
SHAPformer computes numerical Shapley values for a synthetic
data-generating process~\citep{hertel2026shapformer}, which need not match
a fitted model's attributions.

Applications to human action and motion use DeepExplainer with background
samples~\citep{auletta2023predicting,tempel2025choose}.
ShapGCN assesses explanations by perturbing learned edge
weights~\citep{tempel2025explaining}, while ECG analysis uses layer-wise
relevance propagation~\citep{strodthoff2021ecg}.
\section{Datasets}
\label{sec:datasets}

\subsection{Synthetic datasets: designed distributions with oracles}
\label{sec:datasets-synth}

We generate five synthetic datasets with computable marginal and conditional imputations (Appendix~\ref{app:oracle_proofs}). Our five synthetic datasets (Table~\ref{tab:datasets}) vary along three specific
axes:

\textbf{1. Marginal family.}~\texttt{gauss} draws Gaussian coordinates. \texttt{burr} pushes the same latent sequence through a symmetric heavy-tailed Burr copula (tail index $4$), testing whether conclusions survive non-Gaussian marginals.

\textbf{2. Spatial structure.} \texttt{skeleton} uses a joint-correlation matrix inspired by the Human3.6M skeleton  \citep{ionescu2014human36m} ($J{=}17$ joints), giving the strong cross-joint coupling of real motion capture.  \texttt{gauss} and \texttt{burr} use weak coupling ($J{=}5$).

\textbf{3. Temporal structure.}  \texttt{gauss}, \texttt{burr}, and  \texttt{skeleton} use autoregressive temporal correlation. \texttt{gait} and \texttt{skel+gait} use a gait-inspired \emph{periodic} temporal correlation.

Each synthetic dataset defines a three-class classification task. To label a sequence, we average the coordinates of its first joint over all frames, which gives one label per sequence. We then split the sequences into three equally sized classes according to this label. For each dataset we train three classifier architectures. An MLP
sees only the mean of each temporal window, while a 1-D convolutional
network \citep{lecun1998gradient} and a Transformer
\citep{vaswani2017attention} see the raw frames.
Training details are in Appendix~\ref{app:experiment_details}.

\begin{table}[t]
  \centering
  \caption{\textbf{The five synthetic datasets.} All are tensors $\x \in \mathbb{R}^{J\times 3\times T}$. The columns give each dataset's spatial coupling, temporal correlation, and marginal distribution, with the full construction in Appendix~\ref{app:oracle_proofs}. $K$ is the number of temporal windows and $M = J \times K$ the number of joint$\times$window cells. Each dataset has $N{=}200$ evaluation sequences.}
  \label{tab:datasets}
  \footnotesize
  \newcommand{\dsicon}[1]{\raisebox{-0.25ex}{\includegraphics[height=1.7ex]{figures/#1}}}
  \begin{tabular}{clccccccc}
    \toprule
    & Dataset & Spatial coupling $\Sigma_J$ & Temporal $\Sigma_T$ & Marginal & $J$ & $T$ & $K$ & $M$ \\
    \midrule
    \dsicon{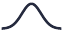}    & \texttt{gauss}  & low            & AR(1)    & Gaussian & $5$  & $16$ & $4$ & $20$ \\
    \dsicon{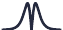}     & \texttt{burr}   & low            & AR(1)    & Burr     & $5$  & $20$ & $5$ & $25$ \\
    \dsicon{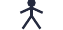}     & \texttt{skeleton}   & high (H3.6M)   & AR(1)    & Gaussian & $17$ & $16$ & $4$ & $68$ \\
    \dsicon{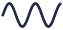}     & \texttt{gait}       & uncoupled      & periodic & Gaussian & $17$ & $16$ & $4$ & $68$ \\
    \dsicon{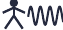} & \texttt{skel+gait}  & high (H3.6M)   & periodic & Gaussian & $17$ & $16$ & $4$ & $68$ \\
    \bottomrule
  \end{tabular}
\end{table}

\subsection{Real-world datasets}
\label{sec:datasets-real}

MotionBench-XAI includes three public datasets spanning spatiotemporal modalities and application domains of particular interest for explanation. \textbf{CARE-PD} \citep{adeli2025multi}: 3-D skeletons of people walking ($J{=}17$). The goal is to predict the patient's Parkinson's severity score. We consider pre-trained architectures (MotionBert \citep{zhu2023motionbert} and MotionAGFormer \citep{mehraban2024motionagformer}), reported as models A and B, over three leave-one-subject-out folds. \textbf{PTB-XL} \citep{wagner2020ptbxl}: recordings of a 12-lead electrocardiogram. The model (a per-fold 1-D ResNet) predicts normal cardiac health or myocardial infarction. \textbf{ESC-50} \citep{piczak2015esc}: 5-second environmental audio clips over 50 classes. We explain the Audio Spectrogram Transformer (AST) \citep{gong2021ast}. Further details are provided in Appendix~\ref{app:real_data_table}.

\section{Results}
\label{sec:results}
\label{sec:salient}

\paragraph{Protocol.} MotionBench-XAI evaluates every method with identical numbers of coalitions at each player set, ensuring identical numbers of classifier evaluations per explanation (Appendix~\ref{app:coalition_protocol}). Every run uses up to $N{=}200$ sequences per fold. Synthetic numbers are averaged over the participating classifier architectures, and $\pm$ is the spread across architectures. Real-data numbers are averaged over the three data folds, and $\pm$ is the spread across folds. A classifier participates if and only if it performs at least $0.05$ above chance and has a train--test gap of at most $0.30$. This criterion excludes the CNN on the two periodic synthetic datasets (overfitting). Per-classifier accuracies are in Appendix~\ref{app:clf_gate}. Synthetic tables
(Tables~\ref{tab:synth-temporal}--\ref{tab:synth-cells}) report $\ECthree$ under game-matched grading (method is scored against the ground truth of its own imputer family). $\ECone$, the mean absolute credit error, and the cross-graded convention are in Appendix~\ref{app:ec1_table}.

\subsection{Synthetic suite: ground-truth grading}
\label{sec:results-synth}

Tables~\ref{tab:synth-temporal}--\ref{tab:synth-cells} score every method on every dataset at the three player sets. The KS-Oracle row runs KernelSHAP with the exact conditional imputer under the same protocol as every other method. The ground truth fills hidden players with the exact conditional \emph{mean}, while KS-Oracle, like every stochastic imputer, averages $R{=}5$ random draws of the hidden players. Its nonzero error therefore measures the cost of the shared protocol itself. Namely, it accounts for the finite draws and, at player sets too large to enumerate every coalition, the shared sampled coalition design (Appendix~\ref{app:oracle_proofs}). Thus the row serves as a reference: error above it reflects the quality of a method's imputer.

\begin{table}[t]
\centering
\caption{\textbf{Synthetic results, temporal players} ($K$ uniform windows). EC3 = one minus the Pearson correlation between estimated and ground-truth credits (lower is better), mean $\pm$ spread across the classifier architectures that perform sufficiently well (Appendix~\ref{app:clf_gate}). Each method is graded against the exact ground truth of its own imputer family (game-matched). Gray text indicates periodic datasets (Result~1); they are excluded from the per-column best. EC1 is reported in Appendix~\ref{app:ec1_table}.}
\label{tab:synth-temporal}
\footnotesize
\setlength{\tabcolsep}{4pt}
\begin{tabular}{@{}llccccc@{}}
\toprule
& Method & \texttt{gauss} & \texttt{skeleton} & \texttt{burr} & \texttt{gait} & \texttt{skel+gait} \\
\midrule

\multirow{5}{*}{\parbox{1.7cm}{\emph{Standard}\\\emph{imputers}}}
& KS-Marginal & 0.36 \msd{0.04} & 0.41 \msd{0.09} & 0.28 \msd{0.04} & \textbf{0.21} \msd{0.08} & \textbf{0.16} \msd{0.07} \\
& WindowSHAP-S & 0.29 \msd{0.06} & 0.50 \msd{0.13} & 0.30 \msd{0.01} & 0.44 \msd{0.04} & 0.54 \msd{0.13} \\
& WindowSHAP-D & 0.59 \msd{0.09} & 0.71 \msd{0.19} & 0.54 \msd{0.12} & 0.86 \msd{0.14} & 0.80 \msd{0.18} \\
& TimeSHAP & 0.35 \msd{0.10} & 0.37 \msd{0.08} & 0.37 \msd{0.15} & 0.43 \msd{0.27} & 0.50 \msd{0.16} \\
& KS-Empirical & 0.51 \msd{0.03} & 0.47 \msd{0.09} & 0.48 \msd{0.03} & 0.31 \msd{0.05} & 0.28 \msd{0.09} \\

\addlinespace

\multirow{3}{*}{\parbox{1.7cm}{\emph{Generative}\\\emph{imputers}}}
& KS-VAEAC & \textbf{0.12} \msd{0.03} & \textbf{0.16} \msd{0.05} & \textbf{0.10} \msd{0.02} & \textcolor{gray}{0.76 \msd{0.08}} & \textcolor{gray}{0.87 \msd{0.04}} \\
& KS-Flow & 0.41 \msd{0.07} & 0.38 \msd{0.09} & 0.29 \msd{0.03} & \textcolor{gray}{0.78 \msd{0.10}} & \textcolor{gray}{0.77 \msd{0.12}} \\
& KS-Gauss (closed form) & 0.24 \msd{0.04} & 0.28 \msd{0.08} & 0.23 \msd{0.03} & \textcolor{gray}{0.69 \msd{0.09}} & \textcolor{gray}{0.61 \msd{0.15}} \\

\midrule

& KS-Oracle (reference) & 0.21 \msd{0.05} & 0.20 \msd{0.06} & 0.19 \msd{0.04} & 0.22 \msd{0.06} & 0.22 \msd{0.03} \\

\bottomrule
\end{tabular}
\end{table}

\begin{table}[t]
\centering
\caption{\textbf{Synthetic results, spatial and spatiotemporal players.} EC3, mean $\pm$ spread across the accuracy-passing classifier architectures, game-matched grading. Gray represents periodic data (Result 1) and is excluded from per-column best.}
\label{tab:synth-spatial}
\label{tab:synth-cells}
\footnotesize
(a) \emph{Spatial players (one player per joint; $J{=}5$ or $17$)}\\[2pt]
\setlength{\tabcolsep}{4pt}
\begin{tabular}{@{}llccccc@{}}
\toprule
& Method & \texttt{gauss} & \texttt{skeleton} & \texttt{burr} & \texttt{gait} & \texttt{skel+gait} \\
\midrule

\multirow{2}{*}{\parbox{1.7cm}{\emph{Standard}\\\emph{imputers}}}
& KS-Marginal & 0.32 \msd{0.03} & 0.20 \msd{0.08} & 0.30 \msd{0.02} & \textbf{0.17} \msd{0.15} & 0.16 \msd{0.09} \\
& KS-Empirical & 0.40 \msd{0.02} & 0.28 \msd{0.05} & 0.41 \msd{0.03} & 0.26 \msd{0.13} & 0.27 \msd{0.09} \\

\addlinespace

\multirow{3}{*}{\parbox{1.7cm}{\emph{Generative}\\\emph{imputers}}}
& KS-VAEAC & \textbf{0.19} \msd{0.06} & \textbf{0.13} \msd{0.02} & \textbf{0.16} \msd{0.06} & 0.19 \msd{0.15} & 0.22 \msd{0.05} \\
& KS-Flow & 0.29 \msd{0.06} & 0.23 \msd{0.06} & 0.35 \msd{0.07} & 0.17 \msd{0.14} & 0.19 \msd{0.07} \\
& KS-Gauss (closed form) & 0.21 \msd{0.07} & 0.17 \msd{0.04} & 0.25 \msd{0.08} & 0.22 \msd{0.15} & \textbf{0.14} \msd{0.07} \\

\midrule

& KS-Oracle (reference) & 0.18 \msd{0.07} & 0.10 \msd{0.04} & 0.19 \msd{0.07} & 0.16 \msd{0.14} & 0.08 \msd{0.04} \\

\bottomrule
\end{tabular}

(b) \emph{Spatiotemporal players (joint$\times$window cells; $M{=}20$--$68$)}\\[2pt]
\setlength{\tabcolsep}{4pt}
\begin{tabular}{@{}llccccc@{}}
\toprule
& Method & \texttt{gauss} & \texttt{skeleton} & \texttt{burr} & \texttt{gait} & \texttt{skel+gait} \\
\midrule

\multirow{2}{*}{\parbox{1.7cm}{\emph{Standard}\\\emph{imputers}}}
& KS-Marginal & 0.19 \msd{0.04} & 0.22 \msd{0.05} & 0.24 \msd{0.01} & \textbf{0.16} \msd{0.12} & \textbf{0.14} \msd{0.09} \\
& KS-Empirical & 0.35 \msd{0.00} & 0.33 \msd{0.02} & 0.39 \msd{0.00} & 0.26 \msd{0.11} & 0.23 \msd{0.10} \\

\addlinespace

\multirow{3}{*}{\parbox{1.7cm}{\emph{Generative}\\\emph{imputers}}}
& KS-VAEAC & 0.20 \msd{0.05} & \textbf{0.22} \msd{0.02} & \textbf{0.24} \msd{0.03} & \textcolor{gray}{0.48 \msd{0.09}} & \textcolor{gray}{0.57 \msd{0.01}} \\
& KS-Flow & 0.26 \msd{0.05} & 0.29 \msd{0.05} & 0.32 \msd{0.03} & \textcolor{gray}{0.40 \msd{0.11}} & \textcolor{gray}{0.53 \msd{0.01}} \\
& KS-Gauss (closed form) & \textbf{0.14} \msd{0.03} & 0.22 \msd{0.03} & 0.24 \msd{0.03} & \textcolor{gray}{0.38 \msd{0.13}} & \textcolor{gray}{0.36 \msd{0.07}} \\

\midrule

& KS-Oracle (reference) & 0.13 \msd{0.03} & 0.18 \msd{0.02} & 0.20 \msd{0.02} & 0.21 \msd{0.08} & 0.22 \msd{0.05} \\

\bottomrule
\end{tabular}
\end{table}

\paragraph{Result 1: Choosing a good player set is the first step towards achieving meaningful explanations.}
On periodic data like \texttt{gait} and \texttt{skel+gait}
(Table~\ref{tab:synth-temporal}), temporal players are unsurprisingly meaningless: they contain the same information and thus receive uniform attribution values. On the other hand, spatial players provide useful insights. For example, on \texttt{skel+gait}, KS-VAEAC's $\ECthree$ falls from $0.87$ for temporal players to $0.22$ for spatial players (Table~\ref{tab:synth-spatial}a). This is a clear-cut example of how correctly choosing players is key in determining explanation usefulness. 

\paragraph{Result 2: Generative imputers are accurate  so long as their conditional models are accurate.}
Under game-matched grading, KS-VAEAC reaches the lowest $\ECthree$ among the displayed methods on every non-periodic dataset at temporal and spatial players (Tables~\ref{tab:synth-temporal} and~\ref{tab:synth-spatial}a). KS-Flow trails it by $0.10$--$0.29$ on the same columns. At joint$\times$window cells, the learned-imputer advantage over KS-Marginal disappears on \texttt{skeleton} and \texttt{burr}: differences are at most $0.005$, below the target variability reported in Appendix~\ref{app:noise_floors}. KS-Gauss has the lowest error on \texttt{gauss} cells and is competitive on \texttt{skeleton} and \texttt{burr} (Table~\ref{tab:synth-cells}b). In summary, the closed-form conditional imputers are just as good as the learned imputers so long as they correctly describe the data.

\paragraph{Result 3: Fine granularity has a measurable evaluation cost.}
At joint$\times$window cells, the grading targets are computed on sampled coalition designs and carry approximation error. We measure this error by recomputing each target using a new set of coalitions. For \texttt{skeleton} cells ($M{=}68$), increasing the target budget from $8{,}192$ to $32{,}768$ coalitions reduces conditional-target disagreement from $\ECthree \approx 0.12$ to $\approx 0.03$ (Appendix~\ref{app:noise_floors}). As such, fine-grained method comparisons must be interpreted alongside this target uncertainty.

\paragraph{Result 4: Grading everything against one ground truth mixes up two different questions.} Prior oracle benchmarks grade every method against the conditional ground truth \citep{aas2021explaining,olsen2022using}, including methods that estimate the marginal game. On our centered data, KS-Zero implements the exact marginal imputation. Its temporal $\ECthree$ on \texttt{gauss} is $0.00$ under game-matched grading but $0.25$ when graded against the conditional target (Appendix~\ref{app:cond_tables}). This error reflects the different targets rather than an inaccurate marginal explanation. The takeaway is the importance of stating the intended game first, then grading against that game's ground truth.

\subsection{Real-world suite: no oracle}
\label{sec:results-real}

Table~\ref{tab:real-faith} evaluates the KernelSHAP family on the three real datasets with the two standard no-oracle metrics, across temporal and spatial player sets. Temporal SHAP variants build in their own temporal players, and KS-Gauss requires a fitted Gaussian model of the full tensor. Hence, both are compared only on the synthetic suite. Joint$\times$window results for CARE-PD and ESC-50 are in Appendix~\ref{app:real_cells}.

\begin{table}[t]
\centering
\caption{\textbf{Real-world results.} Panel (a): faithfulness correlation between summed hidden credits and the prediction drop. Panel (b): PlayerAOPC, the mean drop along the deletion path. Mean $\pm$ spread across data folds; bold: best per column and panel. AOPC magnitudes are comparable within a column.}
\label{tab:real-faith}
\label{tab:real-aopc}
\footnotesize
(a) \emph{Faithfulness correlation ($\uparrow$)}\\[2pt]
\setlength{\tabcolsep}{4pt}
\resizebox{\linewidth}{!}{%
\begin{tabular}{lcccccccc}
\toprule
 & \multicolumn{4}{c}{\emph{Temporal players}} & \multicolumn{4}{c}{\emph{Spatial players}} \\
\cmidrule(lr){2-5}\cmidrule(lr){6-9}
Method & \shortstack{CARE-PD\\{\scriptsize model A, windows}} & \shortstack{CARE-PD\\{\scriptsize model B, windows}} & \shortstack{PTB-XL\\{\scriptsize windows}} & \shortstack{ESC-50\\{\scriptsize windows}} & \shortstack{PTB-XL\\{\scriptsize per lead}} & \shortstack{CARE-PD\\{\scriptsize model A, joints}} & \shortstack{CARE-PD\\{\scriptsize model B, joints}} & \shortstack{ESC-50\\{\scriptsize freq.\ bands}} \\
\midrule
KS-Zero & 0.86 \msd{0.02} & 0.46 \msd{0.05} & 0.88 \msd{0.01} & 0.88 \msd{0.01} & \textbf{0.66} \msd{0.03} & \textbf{0.91} \msd{0.03} & 0.45 \msd{0.02} & 0.78 \msd{0.01} \\
KS-Mean & 0.83 \msd{0.12} & 0.82 \msd{0.08} & 0.88 \msd{0.01} & 0.88 \msd{0.01} & 0.66 \msd{0.03} & 0.76 \msd{0.07} & 0.42 \msd{0.09} & 0.79 \msd{0.01} \\
KS-Marginal & \textbf{0.96} \msd{0.02} & \textbf{0.95} \msd{0.02} & \textbf{0.91} \msd{0.01} & \textbf{0.92} \msd{0.00} & 0.52 \msd{0.01} & 0.81 \msd{0.02} & \textbf{0.56} \msd{0.11} & 0.84 \msd{0.01} \\
\addlinespace
KS-VAEAC & 0.69 \msd{0.05} & 0.73 \msd{0.08} & 0.85 \msd{0.04} & 0.91 \msd{0.00} & 0.55 \msd{0.04} & 0.03 \msd{0.35} & 0.30 \msd{0.06} & \textbf{0.87} \msd{0.01} \\
KS-Flow & 0.66 \msd{0.05} & 0.68 \msd{0.01} & 0.63 \msd{0.02} & 0.90 \msd{0.01} & 0.64 \msd{0.01} & 0.29 \msd{0.17} & 0.45 \msd{0.09} & 0.86 \msd{0.01} \\
\bottomrule
\end{tabular}}
\\[6pt]
(b) \emph{PlayerAOPC ($\uparrow$)}\\[2pt]
\setlength{\tabcolsep}{4pt}
\resizebox{\linewidth}{!}{%
\begin{tabular}{lcccccccc}
\toprule
 & \multicolumn{4}{c}{\emph{Temporal players}} & \multicolumn{4}{c}{\emph{Spatial players}} \\
\cmidrule(lr){2-5}\cmidrule(lr){6-9}
Method & \shortstack{CARE-PD\\{\scriptsize model A, windows}} & \shortstack{CARE-PD\\{\scriptsize model B, windows}} & \shortstack{PTB-XL\\{\scriptsize windows}} & \shortstack{ESC-50\\{\scriptsize windows}} & \shortstack{PTB-XL\\{\scriptsize per lead}} & \shortstack{CARE-PD\\{\scriptsize model A, joints}} & \shortstack{CARE-PD\\{\scriptsize model B, joints}} & \shortstack{ESC-50\\{\scriptsize freq.\ bands}} \\
\midrule
KS-Zero & -0.14 \msd{0.26} & 0.19 \msd{0.06} & 0.19 \msd{0.03} & 0.77 \msd{0.00} & 0.47 \msd{0.04} & -0.08 \msd{0.23} & \textbf{0.43} \msd{0.13} & 0.54 \msd{0.01} \\
KS-Mean & \textbf{0.29} \msd{0.09} & \textbf{0.35} \msd{0.11} & 0.19 \msd{0.03} & 0.78 \msd{0.00} & 0.47 \msd{0.04} & \textbf{0.18} \msd{0.05} & 0.33 \msd{0.09} & 0.55 \msd{0.01} \\
KS-Marginal & 0.12 \msd{0.05} & 0.16 \msd{0.07} & 0.18 \msd{0.01} & \textbf{0.88} \msd{0.01} & 0.37 \msd{0.03} & 0.08 \msd{0.08} & 0.18 \msd{0.02} & \textbf{0.79} \msd{0.02} \\
\addlinespace
KS-VAEAC & 0.09 \msd{0.04} & 0.20 \msd{0.02} & \textbf{0.59} \msd{0.03} & 0.82 \msd{0.00} & 0.45 \msd{0.06} & 0.03 \msd{0.04} & 0.08 \msd{0.01} & 0.69 \msd{0.01} \\
KS-Flow & 0.11 \msd{0.02} & 0.17 \msd{0.01} & 0.27 \msd{0.04} & 0.85 \msd{0.01} & \textbf{0.58} \msd{0.05} & 0.05 \msd{0.02} & 0.13 \msd{0.01} & 0.75 \msd{0.02} \\
\bottomrule
\end{tabular}}
\end{table}

\paragraph{Result 5: Without a ground truth, the two standard metrics disagree, and the best method changes with the dataset.}
On CARE-PD temporal players, faithfulness assigns the highest score to KS-Marginal ($0.96$ and $0.95$ on models A and B; Table~\ref{tab:real-faith}a), while PlayerAOPC ranks KS-Mean highest ($0.29$ and $0.35$; Table~\ref{tab:real-aopc}b). Yet, on ESC-50, KS-Marginal scores highest on both metrics for temporal players. No single method is best across datasets and player sets (Appendix~\ref{app:real_best}). This underscores the importance of comparing possible explanation designs going beyond choice of metric. In view of doing this in practice, we propose a practitioner decision guide in Fig. \ref{fig:decision}. 

\begin{figure}
\centering
\setlength{\fboxsep}{3pt}%
\fbox{\begin{minipage}{\dimexpr\linewidth-2\fboxsep-2\fboxrule\relax}\small
\setlist[enumerate]{topsep=1pt,itemsep=2pt,parsep=2pt,leftmargin=1.4em}%
\textbf{Setting up a Shapley analysis: three questions to answer.}
\begin{enumerate}
\item \textbf{What is known about the data's distribution?}
  If the distribution is known analytically, use the exact conditional
  imputer. If it can be learned from data, train a generative imputer and check that it can reconstruct a hidden player well. If neither holds, use a standard
  imputer (marginal game).
\item \textbf{What is the goal of the explanation?}
  Match the players to the question. Temporal explanations call for temporal players, and so on. More players means better precision (assuming the estimator is good) but also more model evaluations at explanation time.
\item \textbf{How will the explanation be scored?}
  With a ground truth, grade each method against the game its imputer
  plays, recording what that ground truth represents. Without one, report faithfulness and PlayerAOPC together. Only compare
  PlayerAOPC values within one player set.
\end{enumerate}
\end{minipage}}
\captionsetup{justification=raggedright,singlelinecheck=false}
\caption{\textbf{Practitioner decision guide.}
Setup procedure, further described in Appendix~\ref{app:guide}.}
\label{fig:decision}
\end{figure}
\section{Limitations and Conclusions}
\label{sec:discussion}

MotionBench-XAI exposes and compares the design choices behind game-theoretic explanations for spatiotemporal machine learning. The benchmark grades ten attribution methods against exact ground truths on five designed datasets, runs the same pipeline on three real ones, and distills the outcome into useful insights. We conclude with two limitations that offer promising avenues for future work. First, the oracle construction is limited to Gaussian-field and copula families: extending exact ground truths to richer designed manifolds, such as articulated dynamics and discrete events, would widen the benchmark's reach. Second, the suite's dimensions are moderate ($J \leq 17$, $M \leq 68$) and its real-data conclusions are specific to the evaluated checkpoints: scaling to finer granularities and more architectures could better characterize real-world explanation scenarios. Overall, we look forward to MotionBench-XAI powering more intentional and practical use of Shapley explanation in applied spatiotemporal settings.



\subsection*{Acknowledgments}
M. P. and N. M. acknowledge partial support from NSF grant 2602079 and the Noyce Foundation. M. P. and N. M. also thank Goodfire AI for providing access to Silico through its academic research grant program.
M. P. also acknowledges support from NSF CAREER grant 2240158 and the Natural Sciences and Engineering Research Council of Canada (NSERC).

\subsection*{AI use statement}
In this work, we used generative AI tools for the following tasks with required disclosure: designing and giving feedback on experimental methodology, implementing the benchmark methods and evaluation pipeline (including the generators that
sample the synthetic datasets), cleaning and reformatting the real-data caches, assisting in the writing of the proofs of Proposition~1, and assisting in the interpretation of results. Generative AI did not author dataset contents directly: every synthetic sequence is drawn from the released, seeded generators, and all real data come from the cited public
datasets. Translation and qualitative data analysis are not applicable to this work. Additionally, we used generative AI tools for creating and editing software code, editing paper text for readability and accuracy, and identifying
and positioning related literature. We have reviewed all AI-assisted work and code.
Moreover, AI-assisted code is covered by the released test suite. The bulk of this paper's main text was human written while the appendices and supporting details were a combination of human and AI writing. All figures were created by the authors. We take responsibility for the final content of this work, including text, claims, and artifacts produced with the aid of generative AI.


\subsection*{Ethics statement}
All data are public research datasets used under their licenses: CARE-PD (anonymized clinical gait motion capture), PTB-XL (de-identified ECG), ESC-50 (environmental audio), each cited in \S\ref{sec:datasets-real}.  No new human data were collected. We also emphasize that the purpose of the benchmark is to help evaluate and compare explanation methods. It is not intended to produce any clinical recommendations.

\subsection*{Reproducibility statement}
The code release contains the data generators, trainers, attribution and
grading pipelines, the canonical result files behind every table, and
SHA-256-verified reference checkpoints;
Appendix~\ref{app:repro} gives the complete account.  All random streams
behind reported numbers are pinned.

\bibliography{refs}
\bibliographystyle{iclr2027_conference}
\newpage
\appendix

\begin{table}[t]
  \centering
  \caption{\textbf{Notation.}  The four benchmark axes and their symbols.}
  \label{tab:notation}
  \footnotesize
  \renewcommand{\arraystretch}{0.92}
  \begin{tabular}{lll}
    \toprule
    & Symbol & Meaning \\
    \midrule
    \multicolumn{3}{l}{\emph{Data}} \\
    & $\x \in \mathbb{R}^{J\times F\times T}$ & instance: $J$ spatial units, $F$ coords, $T$ frames \\
    & $\clf$ & trained classifier being explained \\
    \addlinespace
    \multicolumn{3}{l}{\emph{Players}} \\
    & $\Pset$, $\np$ & player set (a partition of $\x$), number of players \\
    & $\coal \subseteq \Pset$ & coalition of visible players \\
    \addlinespace
    \multicolumn{3}{l}{\emph{Imputation and game}} \\
    & $\fillx{\coal}$ & $\x$ with players outside $\coal$ imputed \\
    & $\val(\coal) = \clf(\fillx{\coal})$ & value of coalition $\coal$ \\
    & $\shap_i$, $\shaphat$ & Shapley value of player $i$; estimated attribution \\
    \addlinespace
    \multicolumn{3}{l}{\emph{Ground truths and metrics}} \\
    & $\gtstd$, $\gtman$ & exact Shapley values of the marginal / conditional game \\
    & $\ECone$, $\ECthree$ & error in credit sizes / credit rankings (synthetic) \\
    & Faith., PlayerAOPC & deletion-based scores (real data, no oracle) \\
    \bottomrule
  \end{tabular}
\end{table}

\section{Coalition Sampling Protocol}
\label{app:coalition_protocol}

Every attribution in this paper (every method row and every grading
target) is produced by one estimator run on one shared coalition design.
This appendix specifies both.  The tables can then be read with the
guarantee the main text relies on: no difference between two methods at the
same player set can come from the estimation layer.

\paragraph{The estimator.}
Given a design $\mathcal{Z}$ of coalitions, always containing the empty and
full coalitions, every method solves the KernelSHAP weighted least-squares
problem
\begin{equation}
  \bigl(\hat\phi_0, \shaphat\bigr) \;=\; \arg\min_{\phi_0,\,\shap}\;
  \sum_{\coal \in \mathcal{Z}} \pi_{\np}(\coal)\,
  \Bigl[\val(\coal) - \phi_0 - \textstyle\sum_{i \in \coal} \shap_i\Bigr]^2,
  \qquad
  \pi_{\np}(\coal) \;=\; \frac{\np - 1}{\binom{\np}{|\coal|}\,|\coal|\,\bigl(\np - |\coal|\bigr)},
  \label{eq:wls}
\end{equation}
where $\pi_{\np}$ is the Shapley kernel, $\np$ the player count, and
$\phi_0$ the intercept (the value of the empty coalition).  The empty and
full coalitions enter as constraints, implemented as boundary rows with
pinning weight $10^{6}$, so the credits plus intercept sum exactly to the
full-coalition value (the efficiency identity).  The normal
matrix carries a ridge of $10^{-8}$ for numerical stability.

Every value $\val(\coal) = \clf(\fillx{\coal})$ uses the same completion
semantics: the hidden entries of $\fillx{\coal}$ are the \emph{average} of
the imputer's draws, and the classifier is evaluated once on that average
($f$-of-mean semantics, Appendix~\ref{app:oracle_proofs}).  The zero and
mean imputations are deterministic.  Every stochastic imputer averages the
same $R{=}5$ draws per coalition: the imputers that copy from other
sequences (KS-Marginal, KS-Empirical), the conditional models (KS-Gauss,
KS-VAEAC, KS-Flow), and the KS-Oracle reference, which draws from the exact
conditional law.  On the real datasets every imputer uses a single
completion.  KS-Marginal and KS-Empirical copy from the $N{=}200$
evaluation pool, and KS-Empirical's kernel weighting excludes the explained
sequence from its own pool of source sequences.

\paragraph{Exact designs ($\np \leq 12$).}
Whenever $\np \leq 12$, the design enumerates all $2^{\np}$ coalitions and
~\eqref{eq:wls} recovers the Shapley values of the (finite-draw) game
exactly.  This covers every temporal track ($K \in \{4,5\}$ windows), the
spatial tracks of the synthetic datasets with $J{=}5$ joints, the PTB-XL
per-lead track ($\np{=}12$, $4096$ coalitions), and the ESC-50
frequency-band track ($\np{=}4$).

\paragraph{Sampled designs ($\np > 12$).}
Above $12$ players the design is built in two stages.  First, complete size
pairs $(s, \np{-}s)$ are enumerated in kernel-mass order, for as long as the
whole pair fits inside the budget $B$.  The total kernel mass of size $s$ is
proportional to $(\np{-}1)/[s(\np{-}s)]$, so the extreme sizes come first.
Every enumerated row carries its exact kernel weight $\pi_{\np}(\coal)$.

Second, the remaining budget is filled by sampling coalitions without
replacement from the leftover sizes, with size probability proportional to
kernel mass.  Each sampled row carries an importance-corrected uniform
weight (the total leftover kernel mass divided by the number of sampled
rows).  The sampled block is thus a stand-in for the sizes it
represents.  All weights are computed in log space (no binomial-coefficient
overflow at $\np{=}68$), rescaled so the largest non-boundary weight is
$1$, with boundary rows pinned at $10^{6}$.

\paragraph{Shared designs and budgets.}
In the sampled regime, the design for each (dataset, player set, budget) is
drawn once, from seed $7919$, and reused verbatim by every method.  Methods
differ only in the imputer that defines $\val(\coal)$, never in which
coalitions they see.  (Exact designs need no seed: every method sees the
full enumeration.)

The two grading targets are built by the same design generator at their own,
higher budget.  A method's distance to its target therefore includes the
pure design-budget gap, which Appendix~\ref{app:noise_floors} measures
directly.  Method budgets are $B{=}1024$ on the sampled synthetic tracks at
$\np \in \{17, 20, 25\}$ and $B{=}2048$ at $\np{=}68$ and on every sampled
real-data track ($\np \in \{16, 17, 48, 68\}$).  Grading targets are
computed from the same design generator at high budget ($B{=}8192$, raised
to $B{=}32{,}768$ at $\np{=}68$), and exactly (full enumeration) wherever
$\np \leq 12$.  The residual noise of the sampled targets is measured
directly in Appendix~\ref{app:noise_floors}.  Every cell evaluates
$N{=}200$ sequences per dataset and per fold on the real data (CARE-PD's
middle fold holds $180$), three folds pooled.

\paragraph{Metrics on sampled designs.}
On the real data, faithfulness is the Pearson correlation over all $B{+}2$
design rows, including the two boundary rows, mirroring the enumerated
protocol.  The PlayerAOPC deletion-path coalitions (players removed in
decreasing-credit order) are generally not design rows and are evaluated
explicitly.  WindowSHAP and TimeSHAP construct their temporal players
internally (\S\ref{sec:bg-imputation}) and are graded on the same targets as
every other temporal-player method.

\section{Oracle Ground Truths: Statements and Proofs}
\label{app:oracle_proofs}

The synthetic suite is designed so that both estimands of
\S\ref{sec:bg-gt} (the marginal imputation $\mathbb{E}[x_i]$ and the conditional
imputation $\mathbb{E}[x_i \mid \x_{\mathrm{vis}}]$) are available in closed
form.  Every dataset shares one generative structure: a latent Gaussian
field $\bm{z}$ with Kronecker covariance
$\operatorname{Cov}(\operatorname{vec}\bm{z}) = \Sigma_J \otimes I_F
\otimes \Sigma_T$, mapped to observations either directly (Gaussian
marginals) or through a heavy-tailed copula.  Because the designer chooses
$\Sigma_J$, $\Sigma_T$, and the marginal family, both expectations have
closed forms.  This makes both ground truths exactly computable for any trained
classifier.  This appendix states and proves that claim, then connects it to
the grading-target construction used throughout the paper.

\begin{proposition}[Exact ground truths on the synthetic families]
\label{prop:oracle}
Let $\bm{z}$ be the latent Kronecker Gaussian field of a synthetic dataset,
$\operatorname{Cov}(\operatorname{vec}\bm{z}) = \Sigma_J \otimes I_F \otimes
\Sigma_T$, and let the observation $\x$ be either $\bm{z}$ itself (Gaussian
marginals) or its coordinatewise copula transform
$x = \operatorname{sign}(z)\,Q_{\mathrm{Burr}}(2\Phi(|z|){-}1;\,2,2)$ (Burr
marginals).  Fix any trained classifier $\clf$ and any player set $\Pset$
with $\np$ players.  Then:
\begin{enumerate}[label=(\roman*)]
\item the exact marginal imputation is $\mathbb{E}[x_{jft}] = 0$ for every
  coordinate;
\item for every coalition $\coal$, the exact conditional imputation
  $\mathbb{E}[\x_{\mathrm{hid}} \mid \x_{\mathrm{obs}}]$ is available in
  closed form: a linear map of the observed coordinates in the Gaussian case,
  and a one-dimensional integral per hidden coordinate in the Burr case,
  computable to any prescribed accuracy;
\item consequently the two games
  $\val^{\star\mathrm{-mar}}(\coal) = \clf(\x \text{ with hidden entries set
  to } 0)$ and
  $\val^{\star\mathrm{-cond}}(\coal) = \clf(\x \text{ with hidden entries set
  to } \mathbb{E}[\x_{\mathrm{hid}} \mid \x_{\mathrm{obs}}])$ are
  deterministic and exactly evaluable, and for $\np \leq 12$ the ground
  truths $\gtstd$ and $\gtman$ follow from exact enumeration of
  ~\eqref{eq:shapley}.
\end{enumerate}
\end{proposition}

\begin{proof}
\emph{(i)}  The latent field is centered, so in the Gaussian case
$\mathbb{E}[x_{jft}] = 0$ directly.  The symmetric Burr transform is an odd
function of $z$ applied to a symmetric variable, so each transformed marginal
is symmetric about zero and its mean is again exactly zero.  The exact
marginal imputation is therefore the zero imputation on every family.

\emph{(ii)}  Since $\Sigma_F = I_F$, the $F$ coordinate slices are
independent and each carries the covariance $\Sigma = \Sigma_J \otimes
\Sigma_T$ over the $(j,t)$ grid.  For a coalition $\coal$, let $o$ and $h$
index the observed and hidden $(j,t)$ positions.  By standard multivariate
Gaussian conditioning,
\begin{equation*}
  \bm{z}_h \mid \bm{z}_o \;\sim\; \mathcal{N}\!\bigl(
    \Sigma_{ho}\Sigma_{oo}^{-1}\bm{z}_o,\;
    \Sigma_{hh} - \Sigma_{ho}\Sigma_{oo}^{-1}\Sigma_{oh}\bigr),
\end{equation*}
where the Kronecker structure enters only through the efficient construction
of the sub-blocks.  In the Gaussian case $\x = \bm{z}$ and the conditional
mean $W_{\coal}\,\x_{\mathrm{obs}}$ with $W_{\coal} =
\Sigma_{ho}\Sigma_{oo}^{-1}$ is the exact imputation.  $W_{\coal}$ is computed once
per coalition and applied to all $N$ sequences.  In the Burr case the copula
transform is a coordinatewise strictly monotone bijection with analytic
inverse, so the observed coordinates map exactly to latent coordinates and
conditioning happens in the latent field as above.  Each hidden observation
coordinate is then the image $g(z) =
\operatorname{sign}(z)\,Q_{\mathrm{Burr}}(2\Phi(|z|){-}1)$ of a univariate
normal $z \sim \mathcal{N}(\mu, \sigma^2)$, with $\mu, \sigma$ from the
display above.  Its conditional mean is the one-dimensional integral
$\mathbb{E}[x \mid \x_{\mathrm{obs}}] = \int g(\mu + \sigma t)\,
\varphi(t)\,dt$, evaluated by the quadrature below.

\emph{(iii)}  With both imputations deterministic, each game value is one
classifier evaluation, and ~\eqref{eq:shapley} is a finite sum over all
$2^{\np}$ coalitions, feasible for $\np \leq 12$.
\end{proof}

\paragraph{The Burr quadrature.}
The integrand $g(\mu + \sigma t)$ has a square-root cusp where its argument
crosses the copula median $u = \nicefrac{1}{2}$, i.e.\ at $t_0 = -\mu/\sigma$.
This caps any global quadrature rule at an algebraic convergence rate.  We
therefore split the integral at $t_0$ and substitute $t = t_0 \pm y^2$ on
each side.  This removes the cusp and restores the spectral accuracy of
Gauss--Legendre quadrature.  The domain is truncated at $|t| = 13$, where the
excluded normal mass is negligible, and each side uses $64$ nodes ($128$
integrand evaluations per hidden coordinate).  The computed imputations are stable
under node doubling and agree with large-sample Monte-Carlo conditional
means.  The released construction additionally passes a unit-test battery
(unconditional covariance recovery, closed-form conditional moments,
the Shapley efficiency identity, the dummy-player axiom under independence,
Monte-Carlo agreement, and the Burr copula round-trip).

\paragraph{From exact imputations to grading targets.}
The value semantics of the benchmark is $f$-of-mean: a method's
$\val(\coal)$ evaluates the classifier at the \emph{average} of its
completions (Appendix~\ref{app:coalition_protocol}).  As the number of draws
grows, that average converges to the exact conditional (or marginal) mean.
The deterministic games of Proposition~\ref{prop:oracle} are therefore the
infinite-draw limits of the games the methods actually play.  The grading
targets simply plug the limit in directly, pushing the exact imputations through
the trained classifier, so the targets carry no Monte-Carlo noise of their
own.  For $\np \leq 12$ the targets are exact Shapley vectors.  Above that
threshold, the same deterministic value functions are evaluated on the
shared high-budget designs and solved by ~\eqref{eq:wls}, with the
resulting grading error measured in Appendix~\ref{app:noise_floors}.

The KS-Oracle \emph{method} row is distinct from the target: it samples
$R{=}5$ completions from the exact conditional law under the same budget as
every other stochastic imputer.  Its residual against the deterministic target
is therefore finite-draw completion noise plus estimation, and the main
tables use it as a reference.  The reference is not a floor.  The targets
are the infinite-draw means, so a conditional model whose mean is close to
the true one and whose draws are less dispersed carries less completion
noise than exact conditional draws, and can score below the reference
row.

\section{EC1 Tables and the Cross-Graded Convention}
\label{app:ec1_table}
\label{app:cond_tables}

\paragraph{Division of labor between the metrics.}
$\ECthree$ is a per-sequence correlation, hence scale-free and comparable
across datasets and player sets.  The main text reports it exclusively.
$\ECone$ measures the absolute size error of the credits and inherits each
dataset's and game's attribution scale, so it is comparable within a column
but not across columns.  Tables~\ref{tab:app-ec1-temporal},
\ref{tab:app-ec1-spatial}, and~\ref{tab:app-ec1-cells} are the $\ECone$
companions to main-text Tables~\ref{tab:synth-temporal}--\ref{tab:synth-cells},
under the same game-matched grading and the same classifier accuracy
criterion.

\begin{table}[t]
\centering
\caption{\textbf{EC1, temporal players} (companion to Table~\ref{tab:synth-temporal}): mean absolute attribution error against the exact game-matched ground truth, mean $\pm$ spread across the classifier architectures that pass the accuracy criterion (Appendix~\ref{app:clf_gate}). EC1 carries the attribution scale of each dataset, so values compare within a column, not across columns. Gray as in the main table.}
\label{tab:app-ec1-temporal}
\footnotesize
\setlength{\tabcolsep}{4pt}
\begin{tabular}{lccccc}
\toprule
Method & \texttt{gauss} & \texttt{skeleton} & \texttt{burr} & \texttt{gait} & \texttt{skel+gait} \\
\midrule
\multicolumn{6}{l}{\emph{Standard imputers, graded against the marginal ground truth $\gtstd$}} \\
KS-Zero & \textbf{0.000} \msd{0.000} & \textbf{0.000} \msd{0.000} & \textbf{0.000} \msd{0.000} & \textbf{0.000} \msd{0.000} & \textbf{0.000} \msd{0.000} \\
KS-Mean & 0.016 \msd{0.008} & 0.017 \msd{0.011} & 0.009 \msd{0.004} & 0.026 \msd{0.025} & 0.019 \msd{0.001} \\
KS-Marginal & 0.115 \msd{0.052} & 0.105 \msd{0.064} & 0.096 \msd{0.044} & 0.085 \msd{0.081} & 0.102 \msd{0.049} \\
WindowSHAP-S & 0.074 \msd{0.023} & 0.073 \msd{0.037} & 0.073 \msd{0.025} & 0.085 \msd{0.079} & 0.161 \msd{0.010} \\
WindowSHAP-D & 0.099 \msd{0.034} & 0.082 \msd{0.045} & 0.091 \msd{0.031} & 0.111 \msd{0.104} & 0.190 \msd{0.028} \\
TimeSHAP & 0.113 \msd{0.055} & 0.093 \msd{0.057} & 0.106 \msd{0.063} & 0.168 \msd{0.164} & 0.219 \msd{0.109} \\
KS-Empirical & 0.135 \msd{0.052} & 0.116 \msd{0.066} & 0.119 \msd{0.048} & 0.100 \msd{0.095} & 0.124 \msd{0.054} \\
\addlinespace
\multicolumn{6}{l}{\emph{Generative imputers, graded against the conditional ground truth $\gtman$}} \\
KS-VAEAC & 0.060 \msd{0.028} & 0.074 \msd{0.036} & 0.048 \msd{0.021} & \textcolor{gray}{0.102 \msd{0.097}} & \textcolor{gray}{0.157 \msd{0.053}} \\
KS-Flow & 0.128 \msd{0.059} & 0.104 \msd{0.062} & 0.098 \msd{0.044} & \textcolor{gray}{0.101 \msd{0.096}} & \textcolor{gray}{0.130 \msd{0.061}} \\
KS-Gauss (closed form) & 0.092 \msd{0.041} & 0.089 \msd{0.051} & 0.085 \msd{0.038} & \textcolor{gray}{0.093 \msd{0.088}} & \textcolor{gray}{0.100 \msd{0.057}} \\
\midrule
KS-Oracle (reference) & 0.087 \msd{0.039} & 0.080 \msd{0.045} & 0.076 \msd{0.035} & 0.049 \msd{0.047} & 0.056 \msd{0.028} \\
\bottomrule
\end{tabular}
\end{table}

\begin{table}[t]
\centering
\caption{\textbf{EC1, spatial players} (companion to Table~\ref{tab:synth-spatial}), mean $\pm$ spread across the classifier architectures that pass the accuracy criterion (Appendix~\ref{app:clf_gate}), game-matched grading.}
\label{tab:app-ec1-spatial}
\footnotesize
\setlength{\tabcolsep}{4pt}
\begin{tabular}{lccccc}
\toprule
Method & \texttt{gauss} & \texttt{skeleton} & \texttt{burr} & \texttt{gait} & \texttt{skel+gait} \\
\midrule
\multicolumn{6}{l}{\emph{Standard imputers, graded against the marginal ground truth $\gtstd$}} \\
KS-Zero & \textbf{0.000} \msd{0.000} & \textbf{0.001} \msd{0.001} & \textbf{0.000} \msd{0.000} & \textbf{0.002} \msd{0.001} & \textbf{0.002} \msd{0.001} \\
KS-Mean & 0.003 \msd{0.002} & 0.002 \msd{0.001} & 0.002 \msd{0.001} & 0.006 \msd{0.006} & 0.004 \msd{0.001} \\
KS-Marginal & 0.093 \msd{0.042} & 0.030 \msd{0.018} & 0.099 \msd{0.045} & 0.024 \msd{0.023} & 0.027 \msd{0.017} \\
KS-Empirical & 0.127 \msd{0.053} & 0.042 \msd{0.023} & 0.132 \msd{0.058} & 0.034 \msd{0.032} & 0.042 \msd{0.024} \\
\addlinespace
\multicolumn{6}{l}{\emph{Generative imputers, graded against the conditional ground truth $\gtman$}} \\
KS-VAEAC & 0.077 \msd{0.037} & 0.036 \msd{0.019} & 0.064 \msd{0.030} & 0.027 \msd{0.026} & 0.044 \msd{0.021} \\
KS-Flow & 0.116 \msd{0.053} & 0.046 \msd{0.025} & 0.121 \msd{0.057} & 0.024 \msd{0.023} & 0.038 \msd{0.020} \\
KS-Gauss (closed form) & 0.088 \msd{0.044} & 0.040 \msd{0.022} & 0.095 \msd{0.046} & 0.031 \msd{0.029} & 0.035 \msd{0.020} \\
\midrule
KS-Oracle (reference) & 0.081 \msd{0.041} & 0.030 \msd{0.018} & 0.082 \msd{0.041} & 0.023 \msd{0.022} & 0.025 \msd{0.015} \\
\bottomrule
\end{tabular}
\end{table}

\begin{table}[t]
\centering
\caption{\textbf{EC1, spatiotemporal players} (companion to Table~\ref{tab:synth-cells}), mean $\pm$ spread across the classifier architectures that pass the accuracy criterion (Appendix~\ref{app:clf_gate}), game-matched grading. Gray as in the main tables.}
\label{tab:app-ec1-cells}
\footnotesize
\setlength{\tabcolsep}{4pt}
\begin{tabular}{lccccc}
\toprule
Method & \texttt{gauss} & \texttt{skeleton} & \texttt{burr} & \texttt{gait} & \texttt{skel+gait} \\
\midrule
\multicolumn{6}{l}{\emph{Standard imputers, graded against the marginal ground truth $\gtstd$}} \\
KS-Zero & \textbf{0.008} \msd{0.005} & \textbf{0.005} \msd{0.004} & \textbf{0.012} \msd{0.007} & \textbf{0.006} \msd{0.006} & \textbf{0.007} \msd{0.006} \\
KS-Mean & 0.010 \msd{0.006} & 0.006 \msd{0.004} & 0.013 \msd{0.008} & 0.008 \msd{0.008} & 0.008 \msd{0.005} \\
KS-Marginal & 0.028 \msd{0.014} & 0.016 \msd{0.010} & 0.033 \msd{0.016} & 0.013 \msd{0.012} & 0.015 \msd{0.009} \\
KS-Empirical & 0.045 \msd{0.017} & 0.021 \msd{0.012} & 0.046 \msd{0.019} & 0.017 \msd{0.016} & 0.023 \msd{0.010} \\
\addlinespace
\multicolumn{6}{l}{\emph{Generative imputers, graded against the conditional ground truth $\gtman$}} \\
KS-VAEAC & 0.032 \msd{0.015} & 0.019 \msd{0.010} & 0.034 \msd{0.015} & \textcolor{gray}{0.016 \msd{0.015}} & \textcolor{gray}{0.026 \msd{0.008}} \\
KS-Flow & 0.039 \msd{0.017} & 0.021 \msd{0.012} & 0.041 \msd{0.018} & \textcolor{gray}{0.015 \msd{0.014}} & \textcolor{gray}{0.022 \msd{0.009}} \\
KS-Gauss (closed form) & 0.029 \msd{0.012} & 0.019 \msd{0.010} & 0.036 \msd{0.015} & \textcolor{gray}{0.015 \msd{0.014}} & \textcolor{gray}{0.019 \msd{0.009}} \\
\midrule
KS-Oracle (reference) & 0.026 \msd{0.012} & 0.017 \msd{0.009} & 0.032 \msd{0.013} & 0.012 \msd{0.011} & 0.014 \msd{0.008} \\
\bottomrule
\end{tabular}
\end{table}

Two rows need a reading note.  Because every synthetic family is centered,
the exact marginal imputation \emph{is} the zero imputation
(Proposition~\ref{prop:oracle}).  On exact-enumeration tracks, KS-Zero's
value function therefore coincides with the standard target's, and its
$\ECone$ is zero to numerical precision.  KS-Mean's empirical-mean imputation
nearly coincides with it.  These rows restate the grading target rather
than estimate it, which is why the main tables leave them out of the
synthetic ranking.

On sampled tracks the same rows are no longer exactly zero (e.g.\ up to
$0.012$ at cells, Table~\ref{tab:app-ec1-cells}).  The residual is the pure
design gap between the method budget and the high-budget target, a useful
lower bound on what any method can show there.  Conversely, the KS-Oracle
row's nonzero $\ECone$ ($0.076$--$0.087$ on the non-periodic temporal
columns) is its own $R{=}5$ completion noise against the deterministic
target.  The $\ECone$ ordering otherwise mirrors the $\ECthree$ story.

\paragraph{Periodic data and ground-truth credit variation.}
\label{app:periodic_diag}
We can measure the periodic near-null of Result~1 on the ground truth
itself.  The per-sequence spread of the conditional ground-truth credits
drops to $0.05$--$0.06$ on the two periodic datasets, a third of its level
on the other three ($0.12$--$0.15$), while the marginal ground truth keeps
its full spread ($0.14$--$0.23$).  With this little true signal to recover,
the residual error of any imperfect imputer dominates $\ECthree$: the exact
conditional imputer still reaches $0.22$, while every imperfect conditional
imputer sits at $0.61$ or above.  Under spatial players the KS-Oracle
reference itself drops from $0.22$ to $0.08$ on \texttt{skel+gait}: hiding
a joint across all frames restores the credit variation that temporal
windows suppress.

\paragraph{KS-Empirical never improves on KS-Marginal.}
\label{app:empirical}
Across all $15$ dataset $\times$ player-set combinations, KS-Empirical
ranks behind the simpler KS-Marginal, in the main-text $\ECthree$ tables
and in the $\ECone$ tables above.  KS-Empirical approximates a conditional
imputation by copying values from the most similar other sequences in the
evaluation pool, weighted by a kernel (\S\ref{sec:bg-imputation}).  The
copied values still come from other sequences, so they carry the same
off-manifold error as marginal sampling, and the similarity weighting adds
estimation variance on top.

\paragraph{What cross-grading isolates.}
Under \emph{game-matched} grading, each method is scored against the ground
truth of its own imputer family, so its error is estimation error within its
own game.  Under the \emph{cross-graded} convention of prior oracle
benchmarks, every method is scored against the on-manifold ground truth
$\gtman$.  A standard imputer's cross-graded error is then its
estimation error \emph{plus} the divergence $\Delta(\gtstd, \gtman)$ between
the two ground truths.  The two conventions bracket the divergence: for a
method whose game-matched error is near zero (KS-Zero, above), the
cross-graded number is essentially the divergence itself.
Tables~\ref{tab:app-cross-temporal}--\ref{tab:app-cross-cells} report the
cross-graded $\ECthree$ at the three player sets.  The generative-imputer rows
are unchanged from the main tables, since their grading target is already
$\gtman$.

\begin{table}[t]
\centering
\caption{\textbf{Cross-graded EC3, temporal players}: every method graded against the on-manifold (conditional) ground truth, the convention of prior oracle benchmarks. Comparing the standard-imputer block with its game-matched values in Table~\ref{tab:synth-temporal} isolates the ground-truth divergence $\Delta(\gtstd,\gtman)$ from estimator error. Gray columns: on periodic data the conditional ground truth is nearly uniform across windows, so no per-column best is marked.}
\label{tab:app-cross-temporal}
\footnotesize
\setlength{\tabcolsep}{4pt}
\begin{tabular}{lccccc}
\toprule
Method & \texttt{gauss} & \texttt{skeleton} & \texttt{burr} & \texttt{gait} & \texttt{skel+gait} \\
\midrule
\multicolumn{6}{l}{\emph{Standard imputers, graded against the conditional ground truth}} \\
KS-Zero & 0.25 \msd{0.03} & 0.28 \msd{0.09} & 0.23 \msd{0.04} & \textcolor{gray}{0.87 \msd{0.11}} & \textcolor{gray}{0.87 \msd{0.13}} \\
KS-Mean & 0.26 \msd{0.02} & 0.29 \msd{0.09} & 0.23 \msd{0.03} & \textcolor{gray}{0.86 \msd{0.10}} & \textcolor{gray}{0.86 \msd{0.12}} \\
KS-Marginal & 0.45 \msd{0.06} & 0.52 \msd{0.09} & 0.38 \msd{0.03} & \textcolor{gray}{0.86 \msd{0.08}} & \textcolor{gray}{0.84 \msd{0.10}} \\
WindowSHAP-S & 0.46 \msd{0.04} & 0.65 \msd{0.11} & 0.43 \msd{0.04} & \textcolor{gray}{0.90 \msd{0.03}} & \textcolor{gray}{0.91 \msd{0.03}} \\
WindowSHAP-D & 0.68 \msd{0.06} & 0.77 \msd{0.15} & 0.61 \msd{0.09} & \textcolor{gray}{0.94 \msd{0.06}} & \textcolor{gray}{0.98 \msd{0.02}} \\
TimeSHAP & 0.47 \msd{0.06} & 0.48 \msd{0.08} & 0.48 \msd{0.12} & \textcolor{gray}{0.91 \msd{0.05}} & \textcolor{gray}{0.93 \msd{0.03}} \\
KS-Empirical & 0.53 \msd{0.03} & 0.56 \msd{0.09} & 0.51 \msd{0.05} & \textcolor{gray}{0.84 \msd{0.05}} & \textcolor{gray}{0.87 \msd{0.09}} \\
\addlinespace
\multicolumn{6}{l}{\emph{Generative imputers, graded against the conditional ground truth (as in the main tables)}} \\
KS-VAEAC & \textbf{0.12} \msd{0.03} & \textbf{0.16} \msd{0.05} & \textbf{0.10} \msd{0.02} & \textcolor{gray}{0.76 \msd{0.08}} & \textcolor{gray}{0.87 \msd{0.04}} \\
KS-Flow & 0.41 \msd{0.07} & 0.38 \msd{0.09} & 0.29 \msd{0.03} & \textcolor{gray}{0.78 \msd{0.10}} & \textcolor{gray}{0.77 \msd{0.12}} \\
KS-Gauss (closed form) & 0.24 \msd{0.04} & 0.28 \msd{0.08} & 0.23 \msd{0.03} & \textcolor{gray}{0.69 \msd{0.09}} & \textcolor{gray}{0.61 \msd{0.15}} \\
\midrule
KS-Oracle (reference) & 0.21 \msd{0.05} & 0.20 \msd{0.06} & 0.19 \msd{0.04} & \textcolor{gray}{0.22 \msd{0.06}} & \textcolor{gray}{0.22 \msd{0.03}} \\
\bottomrule
\end{tabular}
\end{table}

\begin{table}[t]
\centering
\caption{\textbf{Cross-graded EC3, spatial players}: all methods against the on-manifold ground truth (exact conditional target for $\np \leq 12$, hi-budget sampled target above).}
\label{tab:app-cross-spatial}
\footnotesize
\setlength{\tabcolsep}{4pt}
\begin{tabular}{lccccc}
\toprule
Method & \texttt{gauss} & \texttt{skeleton} & \texttt{burr} & \texttt{gait} & \texttt{skel+gait} \\
\midrule
\multicolumn{6}{l}{\emph{Standard imputers, graded against the conditional ground truth}} \\
KS-Zero & 0.35 \msd{0.19} & 0.31 \msd{0.16} & 0.35 \msd{0.19} & \textbf{0.00} \msd{0.00} & 0.24 \msd{0.19} \\
KS-Mean & 0.35 \msd{0.19} & 0.31 \msd{0.16} & 0.36 \msd{0.19} & 0.05 \msd{0.05} & 0.24 \msd{0.18} \\
KS-Marginal & 0.37 \msd{0.05} & 0.27 \msd{0.09} & 0.39 \msd{0.06} & 0.17 \msd{0.15} & 0.22 \msd{0.13} \\
KS-Empirical & 0.41 \msd{0.02} & 0.29 \msd{0.04} & 0.43 \msd{0.04} & 0.26 \msd{0.13} & 0.28 \msd{0.08} \\
\addlinespace
\multicolumn{6}{l}{\emph{Generative imputers, graded against the conditional ground truth (as in the main tables)}} \\
KS-VAEAC & \textbf{0.19} \msd{0.06} & \textbf{0.13} \msd{0.02} & \textbf{0.16} \msd{0.06} & 0.19 \msd{0.15} & 0.22 \msd{0.05} \\
KS-Flow & 0.29 \msd{0.06} & 0.23 \msd{0.06} & 0.35 \msd{0.07} & 0.17 \msd{0.14} & 0.19 \msd{0.07} \\
KS-Gauss (closed form) & 0.21 \msd{0.07} & 0.17 \msd{0.04} & 0.25 \msd{0.08} & 0.22 \msd{0.15} & \textbf{0.14} \msd{0.07} \\
\midrule
KS-Oracle (reference) & 0.18 \msd{0.07} & 0.10 \msd{0.04} & 0.19 \msd{0.07} & 0.16 \msd{0.14} & 0.08 \msd{0.04} \\
\bottomrule
\end{tabular}
\end{table}

\begin{table}[t]
\centering
\caption{\textbf{Cross-graded EC3, spatiotemporal players}: all methods against the on-manifold ground truth. At $\np{=}68$ (\texttt{skel+gait} cells) the estimating standard imputers' cross-graded error is three to five times their game-matched error (Table~\ref{tab:synth-cells}): at fine granularity the two ground truths assign very different credit across cells.}
\label{tab:app-cross-cells}
\footnotesize
\setlength{\tabcolsep}{4pt}
\begin{tabular}{lccccc}
\toprule
Method & \texttt{gauss} & \texttt{skeleton} & \texttt{burr} & \texttt{gait} & \texttt{skel+gait} \\
\midrule
\multicolumn{6}{l}{\emph{Standard imputers, graded against the conditional ground truth}} \\
KS-Zero & 0.31 \msd{0.10} & 0.31 \msd{0.10} & 0.33 \msd{0.08} & 0.55 \msd{0.05} & 0.72 \msd{0.03} \\
KS-Mean & 0.31 \msd{0.10} & 0.31 \msd{0.10} & 0.33 \msd{0.07} & 0.55 \msd{0.04} & 0.72 \msd{0.03} \\
KS-Marginal & 0.34 \msd{0.06} & 0.35 \msd{0.05} & 0.39 \msd{0.04} & 0.59 \msd{0.06} & 0.71 \msd{0.04} \\
KS-Empirical & 0.39 \msd{0.01} & 0.38 \msd{0.02} & 0.45 \msd{0.02} & 0.58 \msd{0.07} & 0.66 \msd{0.02} \\
\addlinespace
\multicolumn{6}{l}{\emph{Generative imputers, graded against the conditional ground truth (as in the main tables)}} \\
KS-VAEAC & 0.20 \msd{0.05} & \textbf{0.22} \msd{0.02} & \textbf{0.24} \msd{0.03} & 0.48 \msd{0.09} & 0.57 \msd{0.01} \\
KS-Flow & 0.26 \msd{0.05} & 0.29 \msd{0.05} & 0.32 \msd{0.03} & 0.40 \msd{0.11} & 0.53 \msd{0.01} \\
KS-Gauss (closed form) & \textbf{0.14} \msd{0.03} & 0.22 \msd{0.03} & 0.24 \msd{0.03} & \textbf{0.38} \msd{0.13} & \textbf{0.36} \msd{0.07} \\
\midrule
KS-Oracle (reference) & 0.13 \msd{0.03} & 0.18 \msd{0.02} & 0.20 \msd{0.02} & 0.21 \msd{0.08} & 0.22 \msd{0.05} \\
\bottomrule
\end{tabular}
\end{table}

At temporal players, KS-VAEAC is the best cross-graded method on all three
non-periodic datasets ($0.12$/$0.16$/$0.10$).  KS-Zero's definitionally-exact
standard answers rank second ($0.25$/$0.28$/$0.23$): within $0.04$--$0.08$
of the exact conditional imputer itself ($0.21$/$0.20$/$0.19$) and ahead of
every remaining conditional estimator.  A perfect answer to the
\emph{standard} question outranks every genuine estimate of the
\emph{conditional} one but KS-VAEAC's.  That is the quantitative content of
main-text Result~4: cross-graded error is dominated by ground-truth
divergence, not estimation quality.

\paragraph{The divergence grows with granularity.}
At the finest player set the same comparison blows up
(Table~\ref{tab:app-cross-cells}).  On \texttt{skel+gait} cells
($\np{=}68$), the standard imputers score $0.66$--$0.72$ cross-graded.  That
compares with game-matched errors of $0.14$--$0.23$ for the estimating
standard methods (Table~\ref{tab:synth-cells}b) and near-zero $\ECone$ for
the definitional ones (Table~\ref{tab:app-ec1-cells}).  On \texttt{gait}
cells the pattern repeats ($0.55$--$0.59$ cross-graded vs.\ $0.16$--$0.26$
game-matched).  At fine granularity the two ground truths rank cells very
differently, so the choice of game is most consequential exactly where
explanations are most detailed.

\paragraph{Per-classifier breakdown.}
Table~\ref{tab:app-perclf} disaggregates the temporal game-matched
$\ECthree$ by classifier architecture.  The $\pm$ spread in every aggregated
table is the spread across these rows.  Errors are generally lowest under
the MLP and highest under the Transformer (e.g.\ KS-Gauss
$0.20$/$0.23$/$0.29$ across MLP/CNN/Transformer on \texttt{gauss}).  The
MLP's input features are exactly the $K$ window means, which aligns the
classifier's own geometry with the temporal players.  The periodic near-null
affects every admitted architecture individually (KS-VAEAC $0.84$ and
$0.69$ on \texttt{gait} MLP/Transformer), so the gray cells of the main
tables are not an artifact of averaging.  The two $\dagger$ rows are the
excluded CNNs (Appendix~\ref{app:clf_gate}), shown for completeness.
Per-classifier values for the spatial and cell tracks ship with the
canonical result files.

\begin{table}[t]
\centering
\caption{\textbf{Per-classifier EC3, temporal players} (game-matched). The two $\dagger$ rows fail the accuracy criterion (Appendix~\ref{app:clf_gate}) and are excluded from every aggregated table; they are shown here for completeness. Per-classifier values for the spatial and cell player sets ship with the canonical result files in the code release.}
\label{tab:app-perclf}
\scriptsize
\setlength{\tabcolsep}{3.5pt}
\resizebox{\linewidth}{!}{%
\begin{tabular}{llccccccccccc}
\toprule
Dataset & Classifier & \shortstack{KS-Zero} & \shortstack{KS-Mean} & \shortstack{KS-Marginal} & \shortstack{WindowSHAP-S} & \shortstack{WindowSHAP-D} & \shortstack{TimeSHAP} & \shortstack{KS-Empirical} & \shortstack{KS-VAEAC} & \shortstack{KS-Flow} & \shortstack{KS-Gauss} & \shortstack{KS-Oracle} \\
\midrule
\texttt{gauss} & mlp & 0.00 & 0.02 & 0.30 & 0.37 & 0.72 & 0.23 & 0.47 & 0.07 & 0.31 & 0.20 & 0.15 \\
 & cnn & 0.00 & 0.01 & 0.36 & 0.23 & 0.56 & 0.36 & 0.51 & 0.14 & 0.45 & 0.23 & 0.21 \\
 & transformer & 0.00 & 0.02 & 0.41 & 0.27 & 0.50 & 0.47 & 0.55 & 0.14 & 0.47 & 0.29 & 0.27 \\
\addlinespace
\texttt{skeleton} & mlp & 0.00 & 0.02 & 0.29 & 0.68 & 0.98 & 0.26 & 0.35 & 0.10 & 0.26 & 0.19 & 0.13 \\
 & cnn & 0.00 & 0.05 & 0.44 & 0.39 & 0.55 & 0.37 & 0.51 & 0.15 & 0.38 & 0.28 & 0.20 \\
 & transformer & 0.00 & 0.03 & 0.50 & 0.43 & 0.60 & 0.46 & 0.56 & 0.23 & 0.49 & 0.38 & 0.28 \\
\addlinespace
\texttt{burr} & mlp & 0.00 & 0.00 & 0.22 & 0.32 & 0.71 & 0.18 & 0.44 & 0.08 & 0.26 & 0.19 & 0.15 \\
 & cnn & 0.00 & 0.01 & 0.31 & 0.28 & 0.46 & 0.38 & 0.49 & 0.09 & 0.28 & 0.23 & 0.19 \\
 & transformer & 0.00 & 0.03 & 0.31 & 0.30 & 0.45 & 0.55 & 0.51 & 0.13 & 0.32 & 0.28 & 0.24 \\
\addlinespace
\texttt{gait} & mlp & 0.00 & 0.00 & 0.14 & 0.48 & 1.00 & 0.16 & 0.26 & 0.84 & 0.69 & 0.60 & 0.16 \\
 & cnn$^\dagger$ & \textcolor{gray}{0.00} & \textcolor{gray}{0.01} & \textcolor{gray}{0.20} & \textcolor{gray}{0.93} & \textcolor{gray}{0.85} & \textcolor{gray}{0.70} & \textcolor{gray}{0.32} & \textcolor{gray}{0.53} & \textcolor{gray}{0.56} & \textcolor{gray}{0.48} & \textcolor{gray}{0.21} \\
 & transformer & 0.00 & 0.09 & 0.29 & 0.40 & 0.71 & 0.70 & 0.37 & 0.69 & 0.88 & 0.77 & 0.28 \\
\addlinespace
\texttt{skel+gait} & mlp & 0.00 & 0.03 & 0.09 & 0.67 & 0.98 & 0.34 & 0.19 & 0.91 & 0.65 & 0.47 & 0.18 \\
 & cnn$^\dagger$ & \textcolor{gray}{0.00} & \textcolor{gray}{0.01} & \textcolor{gray}{0.32} & \textcolor{gray}{0.82} & \textcolor{gray}{0.89} & \textcolor{gray}{0.67} & \textcolor{gray}{0.38} & \textcolor{gray}{0.77} & \textcolor{gray}{0.68} & \textcolor{gray}{0.53} & \textcolor{gray}{0.29} \\
 & transformer & 0.00 & 0.02 & 0.23 & 0.40 & 0.62 & 0.67 & 0.37 & 0.83 & 0.89 & 0.76 & 0.25 \\
\bottomrule
\end{tabular}}
\end{table}

\section{Grading Error Floors}
\label{app:noise_floors}

Above $\np{=}12$ the grading targets are themselves solved on sampled
designs (Appendix~\ref{app:coalition_protocol}), so the grading itself
carries an approximation error whose size must be measured, not assumed.  We measure it directly.  For each
sampled target we recompute an \emph{independent replicate} target at the
same budget from a disjoint coalition seed.  We then report the error metric
between the replicate and the stored target, averaged over the classifier
architectures.

The floor is the resolution limit of the benchmark at that cell.  A
difference between two methods (or between a method and the target) smaller
than the floor is not readable, and we flag such comparisons rather than
interpret them.  Exact-enumeration cells ($\np \leq 12$) have zero grading
error by construction, so every temporal-player number in the paper is
graded exactly.

\begin{table}[t]
\centering
\caption{\textbf{Grading error floors} for the sampled high-budget targets: EC between the stored grading target and an independent replicate (same budget, disjoint coalition seed), averaged over the gate-passing classifier architectures. Method-vs-target differences below these levels are not interpretable. Exact-enumeration cells ($\np \leq 12$) have zero grading error by construction.}
\label{tab:app-floors}
\footnotesize
\begin{tabular}{llrrcc}
\toprule
Player set & Dataset & $\np$ & Budget & Floor EC3 (cond / marg) & Floor EC1 (cond / marg) \\
\midrule
cells & \texttt{gauss} & 20 & 8,192 & 0.016 / 0.004 & 0.0068 / 0.0030 \\
cells & \texttt{skeleton} & 68 & 8,192 & 0.118 / 0.020 & 0.0128 / 0.0052 \\
cells & \texttt{skeleton} & 68 & 32,768 & 0.026 / 0.003 & 0.0053 / 0.0019 \\
cells & \texttt{burr} & 25 & 8,192 & 0.020 / 0.013 & 0.0082 / 0.0048 \\
cells & \texttt{gait} & 68 & 8,192 & 0.075 / 0.053 & 0.0065 / 0.0060 \\
cells & \texttt{gait} & 68 & 32,768 & 0.014 / 0.022 & 0.0028 / 0.0023 \\
cells & \texttt{skel+gait} & 68 & 8,192 & 0.146 / 0.054 & 0.0114 / 0.0065 \\
cells & \texttt{skel+gait} & 68 & 32,768 & 0.032 / 0.027 & 0.0047 / 0.0026 \\
spatial & \texttt{skeleton} & 17 & 8,192 & 0.006 / 0.003 & 0.0064 / 0.0009 \\
spatial & \texttt{gait} & 17 & 8,192 & 0.001 / 0.001 & 0.0010 / 0.0010 \\
spatial & \texttt{skel+gait} & 17 & 8,192 & 0.005 / 0.001 & 0.0057 / 0.0013 \\
\addlinespace
\multicolumn{6}{l}{Cross-budget agreement at $\np{=}68$ (cond target, $B{=}8{,}192$ vs $B{=}32{,}768$): EC3 0.065} \\
\bottomrule
\end{tabular}
\end{table}

Table~\ref{tab:app-floors} reports the floors.  Two regularities matter for
reading the synthetic tables.  First, the floors grow with the number of
players and are largest for the conditional game at the finest cells.  This
is why the $\np{=}68$ targets are computed at the raised budget
$B{=}32{,}768$ rather than $B{=}8192$ (main-text Result~3).  Second, the
floors are the reading rule for the finest-granularity tables: a gap between
two rows that does not clear the floor of its track is a tie. More specifically, differences below the floor cannot be distinguished from the target's own sampling variability.

The cross-budget agreement row provides an independent check that the two target budgets at
$\np{=}68$ identify the same game.  Method error and evaluation error are thereby kept separate throughout.

\section{Training and Experiment Details}
\label{app:experiment_details}

\paragraph{Synthetic data generation.}
Each dataset is a deterministic function of a fixed seed and the parametric
family of \S\ref{sec:datasets-synth}.  Every dataset materializes $N{=}200$
evaluation sequences.  Classifier training draws $2000$ training and $500$
validation sequences from the same distribution under disjoint seeds, so
train and test never overlap.  All five synthetic datasets share one label
rule, the \emph{proxy} rule of \S\ref{sec:datasets-synth}: class labels are
tertiles of the grand mean of the first spatial component, applied on each
dataset's own draw.  The rule is easy to state, uses no window structure,
and is deliberately hard for classifiers to shortcut, which produces the
wide competence range visible in Appendix~\ref{app:clf_gate}.

\paragraph{Synthetic classifiers.}
For each dataset we train three architectures.  The \emph{MLP} is a 2-layer
network (hidden width $64$, ReLU, BatchNorm on the input layer) consuming
the $K$ window-mean features.  The \emph{CNN} is a 3-layer 1-D convolutional
network (filters $32{\to}32{\to}64$, kernel size $5$, padding $2$, ReLU,
BatchNorm, global average pooling) on raw frames reshaped to
$(J\!\cdot\!F, T)$.  The \emph{Transformer} is encoder-only
($d_{\mathrm{model}}{=}32$, $4$ heads, $2$ layers, feed-forward $128$,
dropout $0.1$, sinusoidal positional encoding, mean-pool readout) on raw
frames reshaped to $(T, J\!\cdot\!F)$.

Training uses Adam with cross-entropy loss for $200$ epochs with early
stopping (patience $20$, best-validation checkpoint kept).  Batch size is
$64$ ($32$ for the Transformer).  Learning rates are $5{\times}10^{-3}$
(MLP), $10^{-3}$ (CNN), and $5{\times}10^{-4}$ (Transformer), and weight
decays $10^{-3}$ (MLP) and $10^{-4}$ (CNN, Transformer).  Runs whose best
validation accuracy falls below $0.50$ are retried with the learning rate
divided by five and the next seed.  Whether a trained classifier's
explanations enter the aggregated tables is decided by the uniform
accuracy criterion of Appendix~\ref{app:clf_gate}.

\paragraph{CARE-PD classifiers.}
We explain the UPDRS-gait severity classifiers released with the CARE-PD
benchmark \citep{adeli2025multi}, one checkpoint per leave-one-subject-out
fold of the BMCLab split.  The classifiers are MotionBERT
\citep{zhu2023motionbert} (DSTformer backbone, $T{=}80$, 3-D H36M
coordinates with crop-scale normalization), MotionAGFormer
\citep{mehraban2024motionagformer} (attention--graph hybrid, clips
zero-padded $80 \to 81$ frames), and POTR \citep{martinez2021pose} (GCN
encoder with a 4-layer Transformer, $d_{\mathrm{model}}{=}128$, root-centered
z-scored coordinates).  The last is excluded by the accuracy criterion
(Appendix~\ref{app:clf_gate}).  Coalition imputations are applied in raw world
coordinates before each backbone's own input normalization.  The explained
output on all real datasets is the model's probability of its own predicted
class on the full sequence: $\val(\coal)$ is the softmax probability of that
class after imputation.

\paragraph{PTB-XL classifier.}
One 1-D residual network \citep{wang2017time,strodthoff2021ecg} is trained
from scratch per fold: three residual stages of filter widths
$64{\to}128{\to}128$ with kernel sizes $\{8,5,3\}$, batch normalization,
global average pooling, and a linear head.  The input is $(12, 1000)$ from
$10$\,s recordings at $100$\,Hz.  The class set is \{Normal, Myocardial
Infarction\}.  Training uses Adam (learning rate $10^{-3}$, weight decay
$10^{-4}$), $30$ epochs, batch size $64$, and gradient clipping at norm
$1$.  The learning-rate schedule is \texttt{ReduceLROnPlateau} (factor
$0.5$, patience $5$, floor $10^{-6}$).  Training also uses a weighted
random sampler for class imbalance.  Folds use the public
\texttt{strat\_fold} column ($1$--$10$).  Fold~$2$ matches the canonical
benchmark split of \citet{strodthoff2021ecg}, and folds~$1$ and~$3$ rotate
the held-out partition:
\begin{center}\small
\begin{tabular}{lccc}
\toprule
Fold index & Train (\texttt{strat\_fold}) & Val & Test \\
\midrule
$1$ & $\{1, \ldots, 7\}$    & $\{9\}$ & $\{8\}$  \\
$2$ & $\{1, \ldots, 8\}$    & $\{9\}$ & $\{10\}$ \\
$3$ & $\{1, \ldots, 7, 9\}$ & $\{8\}$ & $\{10\}$ \\
\bottomrule
\end{tabular}
\end{center}

\paragraph{ESC-50 classifiers.}
One Audio Spectrogram Transformer \citep{gong2021ast} is fine-tuned per
fold from the AudioSet-pretrained base
\texttt{MIT/ast-finetuned-audioset-10-10-0.4593} with a fresh 50-class head. The fold-$f$ model is trained only on fold $f$'s ESC-50 training split ($1600$ clips) and evaluated on the held-out fold ($400$ clips). The best epoch is chosen on the held-out fold. This choice is inconsequential: all runs plateau before the final epoch, and across the
three folds the selected epoch improves held-out accuracy by at most $0.005$ over simply taking the final epoch.

Training uses AdamW (learning rate
$3{\times}10^{-5}$, weight decay $0.01$), $5$ epochs, batch size $16$, a
$5\%$ linear warmup followed by a cosine schedule to zero, gradient clipping
at norm $1$, and seed $42{+}\mathrm{fold}$.  Inputs are log-mel features in
the AST convention ($128$ mel bins $\times$ $1024$ frames), identical to the
attribution-time forward path.  Per-fold held-out accuracies are in
Table~\ref{tab:app-clf-gate}.

\section{The Classifier Accuracy Criterion}
\label{app:clf_gate}

A benchmark of explanations is only as meaningful as the classifiers being
explained.  Every explained classifier must therefore pass one uniform gate
before its results enter any aggregated table: \textbf{(G1)} held-out
accuracy at least $0.05$ above chance, and \textbf{(G2)} a train--test
accuracy gap of at most $0.30$.  G1 removes classifiers that never learned
the task.  G2 removes classifiers whose training-set fit does not transfer,
for which any explanation would describe memorization rather than the
data-generating structure.  Table~\ref{tab:app-clf-gate} reports the full
accuracy accounting for all fifteen synthetic classifiers and the three
real-data classifier families.

\begin{table}[t]
\centering
\caption{\textbf{The classifier accuracy criterion.} A classifier's explanations enter the aggregated tables only if (G1) held-out accuracy $\geq$ chance $+\,0.05$ and (G2) train--test gap $\leq 0.30$. Synthetic: accuracies on the training / validation / evaluation splits. Real: held-out accuracy pooled over the three data folds (per-fold values in parentheses); the training column reports accuracy on the pooled training material of the cross-validation folds. Take-away: one rule yields every exclusion. POTR fails G1, and the \texttt{gait} and \texttt{skel+gait} CNNs fail G2; every other classifier is admitted.}
\label{tab:app-clf-gate}
\footnotesize
\setlength{\tabcolsep}{5pt}
\begin{tabular}{llccccc}
\toprule
Dataset & Classifier & Train & Val.\ & Test & Gap & Admitted \\
\midrule
\multicolumn{7}{l}{\emph{Synthetic (chance $= 1/3$; accuracy bar $0.383$)}} \\
\texttt{gauss} & mlp & 0.644 & 0.662 & 0.590 & +0.054 & \checkmark \\
\texttt{gauss} & cnn & 0.951 & 0.892 & 0.870 & +0.081 & \checkmark \\
\texttt{gauss} & transformer & 0.990 & 0.950 & 0.930 & +0.059 & \checkmark \\
\texttt{skeleton} & mlp & 0.533 & 0.542 & 0.430 & +0.103 & \checkmark \\
\texttt{skeleton} & cnn & 0.961 & 0.878 & 0.865 & +0.096 & \checkmark \\
\texttt{skeleton} & transformer & 0.996 & 0.962 & 0.945 & +0.051 & \checkmark \\
\texttt{burr} & mlp & 0.643 & 0.644 & 0.590 & +0.053 & \checkmark \\
\texttt{burr} & cnn & 0.953 & 0.876 & 0.815 & +0.138 & \checkmark \\
\texttt{burr} & transformer & 0.994 & 0.960 & 0.945 & +0.049 & \checkmark \\
\texttt{gait} & mlp & 0.412 & 0.420 & 0.395 & +0.017 & \checkmark \\
\texttt{gait} & cnn & 0.940 & 0.400 & 0.365 & +0.575 & \ding{55}\ (G2) \\
\texttt{gait} & transformer & 0.993 & 0.846 & 0.845 & +0.148 & \checkmark \\
\texttt{skel+gait} & mlp & 0.512 & 0.544 & 0.515 & -0.003 & \checkmark \\
\texttt{skel+gait} & cnn & 1.000 & 0.520 & 0.500 & +0.500 & \ding{55}\ (G2) \\
\texttt{skel+gait} & transformer & 0.969 & 0.898 & 0.825 & +0.143 & \checkmark \\
\addlinespace
\multicolumn{7}{l}{\emph{Real (chance: CARE-PD $1/3$, PTB-XL $1/2$, ESC-50 $1/50$)}} \\
CARE-PD & MotionBERT & 0.974 & --- & 0.698 {\scriptsize(0.24/1.00/0.97)} & +0.276 & \checkmark \\
CARE-PD & MotionAGFormer & 0.983 & --- & 0.726 {\scriptsize(0.31/1.00/0.97)} & +0.258 & \checkmark \\
CARE-PD & POTR & --- & --- & 0.374 & --- & \ding{55}\ (G1) \\
PTB-XL & ResNet1d & 0.968 & --- & 0.952 {\scriptsize(0.95/0.95/0.94)} & +0.016 & \checkmark \\
ESC-50 & AST (3 fold models) & 0.999 & --- & 0.965 {\scriptsize(0.95/0.98/0.96)} & +0.034 & \checkmark \\
\bottomrule
\end{tabular}
\end{table}

One rule yields every exclusion in the paper.  POTR on CARE-PD fails G1
(pooled held-out accuracy $0.374$ against the three-class bar of $0.383$).
The CNNs on the two periodic synthetic datasets fail G2 (\texttt{gait}
$+0.575$, \texttt{skel+gait} $+0.500$).  Their near-perfect training
accuracy collapses to near-chance held out, the signature of memorizing
periodic templates.  Every other classifier passes, including both CARE-PD
backbones (gaps $+0.276$ and $+0.258$), the PTB-XL ResNets ($+0.016$), and
the per-fold ESC-50 ASTs ($+0.034$).

The gate is an aggregation rule for the synthetic suite, not a
data-deletion rule.  The two excluded CNNs' per-cell results are still
computed and still ship in the canonical result files (and appear grayed in
Table~\ref{tab:app-perclf}).  They are only barred from the aggregated
tables and from ranking claims.  POTR fails the accuracy condition, so no
attributions are computed for it (explaining a classifier that has not
learned the task grades noise) and only its gate record ships.  Synthetic
aggregates on \texttt{gait} and \texttt{skel+gait} therefore average the
MLP and Transformer.  All other synthetic aggregates average three
architectures.  CARE-PD results are reported for the two passing backbones
as models A and B.

\section{Real-Data Protocol and Additional Tables}
\label{app:real_data_table}
\label{app:real_cells}

Table~\ref{tab:app-real-protocol} collects the full real-data protocol.
Each dataset contributes three tracks (temporal windows, spatial units, and
their intersection cells) under the shared estimation layer of
Appendix~\ref{app:coalition_protocol}: exact enumeration wherever
$\np \leq 12$, and shared sampled designs at $B{=}2048$ above.  Each track
evaluates $N{=}200$ sequences per fold ($200/180/200$ on CARE-PD's
subject-wise folds), three folds pooled, with the $\pm$ spread across
folds.  All classifiers are per-fold models that pass the gate of
Appendix~\ref{app:clf_gate}.  POTR is excluded there and appears in no
result table.

\begin{table}[t]
\centering
\caption{\textbf{Real-data protocol.}  Tasks, classifiers, and the three
player-set tracks per dataset.  All tracks: $3$ folds, $200$ sequences per
fold ($200/180/200$ on CARE-PD), folds pooled.  Sampled designs share the
coalition protocol of Appendix~\ref{app:coalition_protocol} (seed $7919$).
CARE-PD folds are leave-one-subject-out.  ESC-50 uses folds $1$--$3$ of its
$5$ stratified folds.}
\label{tab:app-real-protocol}
\small
\setlength{\tabcolsep}{4pt}
\resizebox{\linewidth}{!}{%
\begin{tabular}{lllllc}
\toprule
Dataset & Task & Classifiers (one per fold) & Track & $\np$ & Coalitions \\
\midrule
CARE-PD & UPDRS-gait severity (3-class) & MotionBERT (A), MotionAGFormer (B); & windows ($K{=}4$) & $4$ & exact ($2^{4}$) \\
 & $17$ H36M joints, $T{=}80$ & POTR excluded (Appendix~\ref{app:clf_gate}) & joints & $17$ & sampled, $B{=}2048$ \\
 & & & joint$\times$window cells & $68$ & sampled, $B{=}2048$ \\
\addlinespace
PTB-XL & Normal vs.\ MI (binary) & ResNet1d & windows ($K{=}4$) & $4$ & exact ($2^{4}$) \\
 & $12$ leads, $T{=}1000$ ($100$\,Hz) & & leads & $12$ & exact ($2^{12}$) \\
 & & & lead$\times$window cells & $48$ & sampled, $B{=}2048$ \\
\addlinespace
ESC-50 & environmental audio (50-class) & AST fine-tune & windows ($K{=}4$) & $4$ & exact ($2^{4}$) \\
 & $128$ mel bins, $T{=}1024$ & & freq.\ bands (mel quartiles) & $4$ & exact ($2^{4}$) \\
 & & & band$\times$window cells & $16$ & sampled, $B{=}2048$ \\
\bottomrule
\end{tabular}}
\end{table}

\paragraph{Best methods per dataset and player set.}
\label{app:real_best}
On CARE-PD, standard imputers score highest at every granularity and under
both metrics (windows, joints, and cells; Table~\ref{tab:real-faith}a and
Table~\ref{tab:app-real-cells}).  On ESC-50, KS-Marginal scores highest on
both metrics at windows ($0.92$ faithfulness, $0.88$ PlayerAOPC), but on
the frequency-band and band$\times$window players the two metrics
separate: KS-VAEAC is best on faithfulness ($0.87$, $0.88$) and KS-Marginal
on PlayerAOPC ($0.79$, $0.85$).  On PTB-XL no method is best everywhere.
KS-Marginal has the highest faithfulness at windows.  Per lead, KS-Zero and
KS-Mean have the highest faithfulness ($0.66$) with KS-Flow close behind
($0.64$), and KS-Flow has the clearly highest PlayerAOPC ($0.58$ vs.\
$0.47$).

\paragraph{Spatiotemporal cells on real data.}
Table~\ref{tab:app-real-cells} extends the main-text real-data tables to the
finest player set: joint$\times$window cells on CARE-PD ($\np{=}68$),
lead$\times$window cells on PTB-XL ($\np{=}48$), and band$\times$window
cells on ESC-50 ($\np{=}16$).

\begin{table}[t]
\centering
\caption{\textbf{Real-world results, spatiotemporal players} (joint$\times$window / lead$\times$window / band$\times$window cells). Faithfulness ($\uparrow$) and PlayerAOPC ($\uparrow$), pooled over the three data folds, $\pm$ spread across folds; bold: best per column and metric. Protocol identical to Table~\ref{tab:real-faith} (coalition budget 2048, $N{=}200$ per fold).}
\label{tab:app-real-cells}
\footnotesize
\setlength{\tabcolsep}{4pt}
\resizebox{\linewidth}{!}{%
\begin{tabular}{lcccccccc}
\toprule
 & \multicolumn{4}{c}{\emph{Faithfulness}} & \multicolumn{4}{c}{\emph{PlayerAOPC}} \\
\cmidrule(lr){2-5}\cmidrule(lr){6-9}
Method & \shortstack{CARE-PD A\\{\scriptsize $\np{=}68$}} & \shortstack{CARE-PD B\\{\scriptsize $\np{=}68$}} & \shortstack{PTB-XL\\{\scriptsize $\np{=}48$}} & \shortstack{ESC-50\\{\scriptsize $\np{=}16$}} & \shortstack{CARE-PD A\\{\scriptsize $\np{=}68$}} & \shortstack{CARE-PD B\\{\scriptsize $\np{=}68$}} & \shortstack{PTB-XL\\{\scriptsize $\np{=}48$}} & \shortstack{ESC-50\\{\scriptsize $\np{=}16$}} \\
\midrule
KS-Zero & \textbf{0.89} \msd{0.03} & 0.66 \msd{0.23} & 0.74 \msd{0.03} & 0.77 \msd{0.01} & -0.11 \msd{0.25} & \textbf{0.39} \msd{0.13} & \textbf{0.39} \msd{0.04} & 0.72 \msd{0.00} \\
KS-Mean & 0.82 \msd{0.05} & \textbf{0.68} \msd{0.15} & 0.74 \msd{0.03} & 0.78 \msd{0.01} & \textbf{0.19} \msd{0.04} & 0.33 \msd{0.09} & 0.39 \msd{0.04} & 0.72 \msd{0.00} \\
KS-Marginal & 0.79 \msd{0.03} & 0.50 \msd{0.03} & 0.57 \msd{0.02} & 0.87 \msd{0.01} & 0.08 \msd{0.08} & 0.19 \msd{0.01} & 0.32 \msd{0.02} & \textbf{0.85} \msd{0.01} \\
\addlinespace
KS-VAEAC & 0.39 \msd{0.24} & 0.54 \msd{0.10} & \textbf{0.78} \msd{0.03} & \textbf{0.88} \msd{0.01} & 0.02 \msd{0.04} & 0.08 \msd{0.01} & 0.30 \msd{0.05} & 0.80 \msd{0.01} \\
KS-Flow & 0.42 \msd{0.16} & 0.52 \msd{0.07} & 0.61 \msd{0.05} & 0.86 \msd{0.01} & 0.05 \msd{0.03} & 0.12 \msd{0.01} & 0.26 \msd{0.03} & 0.81 \msd{0.01} \\
\bottomrule
\end{tabular}}
\end{table}

The main-text findings persist at this granularity.  CARE-PD keeps its
highest-scoring standard imputers under both metrics (Result~5), and its two
backbones continue to disagree on which standard imputer scores highest.  The metric
disagreement of Result~5 is at its starkest on model A, where KS-Zero has
the best faithfulness ($0.89$) and a negative PlayerAOPC.  ESC-50
reproduces the split of Result~5 at cells: KS-VAEAC leads faithfulness
($0.88$) while KS-Marginal leads PlayerAOPC ($0.85$).  PTB-XL remains the
contested case and reproduces the metric split at cells: KS-VAEAC leads
faithfulness ($0.78$) while KS-Zero and KS-Mean lead PlayerAOPC ($0.39$),
with KS-Marginal behind on both.

Reading across Tables~\ref{tab:real-faith}
and~\ref{tab:app-real-cells}, no method family wins PTB-XL under both
metrics at any granularity.  This is the strongest real-data illustration
that a no-oracle metric is itself a declared choice (Result~5).  As
everywhere, AOPC magnitudes shrink with player granularity and are only
comparable within one player set (\S\ref{sec:bg-metrics}).

\section{Imputer Details and Capability Checks}
\label{app:imputer_detail}

\paragraph{KS-Gauss (closed-form Gaussian conditional).}
Following the parametric approach of \citet{aas2021explaining}, KS-Gauss
fits one joint Gaussian to the training pool with Ledoit--Wolf-shrunk
covariance.  For each coalition, it fills the hidden coordinates from the
closed-form conditional distribution given the visible ones ($R{=}5$ draws,
averaged).  It has no learned components.  Its only assumption is Gaussian
dependence, which is exactly satisfied on the Gaussian-marginal synthetic
families and deliberately violated on \texttt{burr}.

\paragraph{KS-VAEAC.}
Following the arbitrary-conditioning framework of
\citet{ivanov2018variational,olsen2022using}, the VAEAC imputer is a
per-frame-token transformer with three subnetworks (full encoder, prior
encoder, decoder).  Each is a 2-layer transformer trunk with
$d_{\mathrm{model}}{=}256$, $8$ heads, feed-forward $512$, dropout $0.1$,
and sinusoidal frame positional encoding.  Each frame is one token (full
encoder sees $[\x, \mathbf{m}]$, prior encoder $[\x \odot \mathbf{m},
\mathbf{m}]$, decoder $[\mathbf{z}, \x \odot \mathbf{m}, \mathbf{m}]$ for
mask $\mathbf{m}$).  The per-frame latent dimension is $64$, and a Gaussian
output head carries a single learnable global $\log\sigma$.

The loss is negative log-likelihood on hidden entries plus
$\mathrm{KL}(q\,\|\,p)$, optimized with Adam (learning rate $10^{-3}$,
gradient clip $1.0$, batch $16$) under randomly mixed masks
(temporal-window, spatial-unit, and element-wise).  Training runs for $80$
epochs on the synthetic families (the motion-capture imputers train through
the external CARE-PD repository's configurations).  Every released
checkpoint records its full training configuration.  At inference the
latent is drawn from the prior encoder, decoded, and the observed entries
are restored exactly.  In the notation of \S\ref{sec:bg-imputation}, the
imputer draws a latent $z$ from the amortized conditional prior $p_\psi$
and decodes it with $g_\theta$:
\begin{equation}
  \tilde{\x}_{\bar{\coal}}^{\textsc{vaeac}} = g_\theta(z, \x_{\coal}),
  \qquad z \sim p_\psi(z \mid \x_{\coal}).
  \label{eq:generative}
\end{equation}

\paragraph{KS-Flow.}
KS-Flow is a conditional optimal-transport flow-matching imputer
\citep{lipman2023flow,tong2024improving}.  The velocity network is an
encoder-only transformer ($d{=}256$, $4$ heads, $4$ layers, GELU) with
frame-level tokenization.  It is trained to minimize
$\mathbb{E}\,\lVert v_\theta(\x_t, t) - (\x_1 - \x_0)\rVert^2$ along the
interpolant $\x_t = (1{-}t)\x_0 + t\x_1$ with
$\x_0 \sim \mathcal{N}(\bm{0}, I)$.  At inference, $20$ midpoint-Euler ODE
steps are interleaved with RePaint harmonization
\citep{lugmayr2022repaint}: at each step the observed coordinates are
snapped to their interpolant values, so the trajectory conditions on the
visible data.  In the notation of \S\ref{sec:bg-imputation}, the imputer
transports noise to the manifold along the learned velocity field
$v_\eta$:
\begin{equation}
  \tilde{\x}_{\bar{\coal}}^{\textsc{flow}} = \x_{\bar{\coal}}^{(1)},
  \qquad \tfrac{d\x_{\bar{\coal}}^{(t)}}{dt} =
  v_\eta\bigl(\x_{\bar{\coal}}^{(t)}, t,  \x_{\coal}\bigr),
  \qquad \x_{\bar{\coal}}^{(0)} \sim \mathcal{N}(0, I).
\end{equation}

\paragraph{Real-data checkpoints.}
One VAEAC and one flow imputer are trained per real dataset on that
dataset's training pool and reused across all of its tracks and folds.  The
PTB-XL pair uses a smaller configuration recorded in its released
checkpoints: $d_{\mathrm{model}}{=}128$, $4$ heads, $4$ layers, feed-forward
$256$, VAEAC latent $32$, and flow with $20$ ODE steps.  Every released
checkpoint stores its full architecture dictionary, and the loaders
instantiate from it.

\paragraph{The capability check.}
A generative imputer is a \emph{model} of the conditional distribution, and how
good a model it is depends on what is being hidden.  We therefore
characterize every learned imputer with a hide-one-reconstruct check, run
per player-set mask family.  The check hides one player (a spatial unit, a
temporal window, or a cell), imputes its conditional mean from the visible
rest ($8$ draws averaged), and correlates the imputation with the hidden
ground truth.  Scores are computed over $3$ folds and $24$ sequences per
fold.

Two controls calibrate the score.  The \emph{donor} control copies the
hidden entries from the most-correlated other unit of the same sequence:
the recovery available from raw redundancy alone, with no model.  The
\emph{unconditional} control runs the same imputer with everything hidden:
the recovery available from dataset stereotypy alone, with no conditioning.
An imputer demonstrably conditions on the visible context exactly when its
recovery clears both controls.  Table~\ref{tab:app-imputer-gate} reports
the full matrix.

\begin{table}[t]
\centering
\caption{\textbf{Imputer capability check}: hide-one-recover correlation between the imputed conditional mean and the hidden ground truth, per player-set mask family (mean $\pm$ spread over masks; 3 folds, 24 sequences per fold, 8 draws). \emph{Donor}: copy the hidden entries from the most-correlated other unit of the same sequence -- the recovery available from raw redundancy alone. \emph{Unconditional}: imputer run with everything hidden. A learned imputer conditions on the visible context exactly when its recovery clears both controls.}
\label{tab:app-imputer-gate}
\footnotesize
\begin{tabular}{llccc}
\toprule
Dataset & Imputer & Spatial & Temporal & Cell \\
\midrule
CARE-PD & VAEAC & 1.00 \msd{0.00} & 0.96 \msd{0.12} & 1.00 \msd{0.00} \\
 & Flow & 1.00 \msd{0.00} & 1.00 \msd{0.01} & 1.00 \msd{0.00} \\
 & Donor control & 0.97 & 0.98 & 0.98 \\
 & Unconditional & \multicolumn{3}{c}{0.99 (all families)} \\
\addlinespace
PTB-XL & VAEAC & 0.96 \msd{0.09} & 0.05 \msd{0.10} & 0.98 \msd{0.07} \\
 & Flow & 0.95 \msd{0.12} & 0.10 \msd{0.15} & 0.99 \msd{0.04} \\
 & Donor control & -0.01 & -0.01 & -0.01 \\
 & Unconditional & \multicolumn{3}{c}{0.00 (all families)} \\
\addlinespace
ESC-50 & VAEAC & 0.87 \msd{0.24} & 0.31 \msd{0.35} & 0.36 \msd{0.39} \\
 & Flow & 0.88 \msd{0.17} & 0.29 \msd{0.33} & 0.37 \msd{0.39} \\
 & Donor control & 0.62 & 0.16 & 0.12 \\
 & Unconditional & \multicolumn{3}{c}{0.75 (all families)} \\
\bottomrule
\end{tabular}
\end{table}

\paragraph{PTB-XL: conditioning proven cross-lead, absent cross-window.}
The twelve leads observe one cardiac source, and the imputers exploit it.
A hidden lead is recovered at $0.95$--$0.96$ correlation, against a donor
control of $-0.01$ and an unconditional control of $0.00$.  That is genuine
cross-lead conditioning, not redundancy or stereotypy.  Cell masks behave
the same way ($0.98$--$0.99$), since a hidden lead--window cell is
constrained by both its lead's other windows and the other leads.

A hidden \emph{whole window}, by contrast, removes that time span from
every lead at once, leaving nothing contemporaneous to condition on.
Recovery drops to $0.05$--$0.10$: ECG imputation interpolates across leads,
not across time.  The same imputer is thus a trustworthy conditional model
under spatial and cell players and not under temporal players.  The
near-perfect lead recovery also explains the flat per-lead attributions of
\S\ref{sec:results-real}: deleting a lead removes almost no information, so
low per-lead credit is the correct conditional answer.

\paragraph{CARE-PD: at the redundancy ceiling.}
Gait is so stereotyped and cross-joint-redundant that every number in the
CARE-PD block saturates.  The imputers recover hidden players at
$0.96$--$1.00$, but the donor control already reaches $0.97$--$0.98$ and
the unconditional control $0.99$.  Recovery itself is excellent (the
imputations are realistic), but the check cannot separate conditioning from dataset
stereotypy where stereotypy alone suffices.  On such data the two are
observationally equivalent, and the controls say so explicitly.

\paragraph{ESC-50: conditioning proven cross-band, partial elsewhere.}
On spatial (frequency-band) masks the imputers clear both controls
($0.87$--$0.88$ against donor $0.62$ and unconditional $0.75$): hidden bands
are genuinely reconstructed from the visible spectrum.  On temporal and cell
masks, recovery ($0.29$--$0.37$) clears the donor control ($0.16$/$0.12$)
but not the unconditional one: what is recovered there reflects
dataset-level spectrogram structure more than clip-specific content.

Read as a whole, the check is a capability map, not a verdict.  Before any
attribution is computed, it tells a practitioner which mask families (and
hence which player sets) a given generative imputer is a trustworthy
conditional model for.  It also supplies the controls needed to say so on a
new dataset.  This is the measurement behind the decision guide's
``generative imputer \emph{with} a capability check'' branch
(Appendix~\ref{app:guide}).

\paragraph{Practical interpretation.}
If the data distribution is analytically known, the exact conditional
imputer, or its closed-form Gaussian approximation, is the natural choice:
KS-Gauss stays competitive even when its Gaussian assumption is violated
(Result~2).  Where the check finds near-perfect reconstruction, as for the
hidden ECG lead above, hiding a player removes almost no information, so
near-uniform credit across players is the correct explanation rather than
a failure.  Where no conditional model passes the check, standard imputers
are the sound choice, interpreted as answers to the marginal game: an
exact answer to the marginal question is more useful than an inaccurate
answer to the conditional one (Result~4).

\section{Reproducibility}
\label{app:repro}

The release contains the complete pipeline: the synthetic data generators,
classifier and imputer trainers, and the attribution and grading pipelines
(coalition designs, imputers, WLS solver, exact and high-budget targets,
capability check).  It also contains every configuration file behind every
reported cell.  The canonical per-cell result files ship with the code,
with a table-by-table map from each paper table to its file and rerun entry
point, so every reported number is verifiable in minutes, without a GPU.

Reference checkpoints are hosted with a SHA-256 manifest, verified on
download, and split by the source datasets' licenses.  Weights whose source
license forbids redistributing derivatives retrain from the documented
recipes.  Each retraining path was validated by regenerating the shipped
reference tracks bit-exactly.  The packaged estimator, oracle imputations, and
player partitions are pinned by the released test suite, and the oracle
construction carries the unit-test battery of
Appendix~\ref{app:oracle_proofs}.  Per-cell results for the gate-excluded
synthetic CNNs are included (Appendix~\ref{app:clf_gate}), so the criterion is
reproducible and revisable.  The random streams behind every reported
number are pinned in the configurations and entry points: data generation,
training, sampled coalition designs (seed $7919$), imputer draws, and
evaluation subsets.

The CARE-PD backbones and imputers fine-tune through the external CARE-PD
repository.  The WindowSHAP baselines wrap a third-party package.  Display
names map to configuration names (\texttt{gauss} $\to$
\texttt{gaussian\_k4}, \texttt{burr} $\to$ \texttt{burr\_m5},
\texttt{skeleton} $\to$ \texttt{skeleton\_structured},
\texttt{gait} $\to$ \texttt{gait\_periodic}, \texttt{skel+gait} $\to$
\texttt{skeleton\_gait\_combined}).  KS-Gauss appears in the canonical
files as \texttt{ks\_shapr}, and WindowSHAP-S/D as
\texttt{windowshap\_stationary}/\texttt{windowshap\_dynamic}.

Experiments ran on a shared SLURM-managed GPU cluster.  Every training,
attribution, and grading job fits on a single GPU, and the benchmark
imposes no multi-node requirements.

\section{Additional Discussion of Explanation Evaluation}
\label{app:eval_critique}

Existing evaluation practice can certify explanations for the wrong
reasons.
Saliency maps can be insensitive to model and data
randomization~\citep{adebayo2018sanity}. Deletion-style evaluations
conflate attribution quality with the distribution shift they
induce~\citep{hooker2019roar}, and faithfulness metrics inherit related
instabilities~\citep{bhatt2020evaluating,hedstrom2023quantus}. Our
metric-disagreement finding (Result~5) extends this lineage to imputer-based
Shapley evaluation on real spatiotemporal data.

\section{Setting up a Shapley Analysis in Practice}
\label{app:guide}

We summarize MotionBench-XAI's findings into a decision guide
(Figure~\ref{fig:decision}) organized by the three questions a practitioner
should ask before computing Shapley attributions on spatiotemporal data.
Each question selects one of the choices the benchmark measures, and each
answer is a choice to state alongside the explanation.

\paragraph{What is known about the data's distribution?}
This question selects the imputer. If the distribution can be learned, a
generative imputer is appropriate, but it should not be trusted blindly.
Run the capability check first: hide one player, ask the imputer to
reconstruct it from the rest, and compare the reconstruction against simply
copying the most similar visible unit. The comparison against copying
matters, because redundant data makes any imputer look capable.
Appendix~\ref{app:imputer_detail} gives the full check and its
interpretation.

\paragraph{What is the goal of the explanation?}
This question selects the players, and the players should match the
question, not the attribution method. To ask which phase of a movement
matters, use temporal windows. To ask which anatomical unit matters, use
spatial players. To localize in both space and time, use
joint$\times$window cells, and budget for them: cells need more classifier
evaluations, and the grading error floor of Result~3 should be measured and
reported next to the results. Two warnings transfer directly from the
synthetic suite. On periodic data, temporal windows can make the
conditional game unanswerable (Result~1). And existing temporal SHAP
methods hard-code time windows as their players, so changing the question
means changing the method (\S\ref{sec:bg-imputation}).

\paragraph{How will the explanation be scored?}
With a ground truth, grade each method against the ground truth of its own
game, and state which game that is (Result~4). Without one, report
faithfulness \emph{and} PlayerAOPC together, and expect them to disagree
(Result~5). The best method under one metric is a statement about how that
metric is built, not about explanation quality in general. And AOPC values
should never be compared across player granularities
(\S\ref{sec:bg-metrics}).


\end{document}